\documentclass[cameraready]{template/mac_automl_xmu_blue}

\makeatletter
\def\input@path{{content/}}
\makeatother
\graphicspath{{content/}}

\usepackage[toc,page,header]{appendix}

\usepackage{xargs}
\usepackage{todonotes}
\usepackage{tabularx}
\usepackage{booktabs}
\usepackage{float}
\providecommand{\Description}[1]{}

\setleftheadercontent{%
  \headerlogospace{1.6mm}%
  \headerlogo[-0.08\height]{15.2mm}{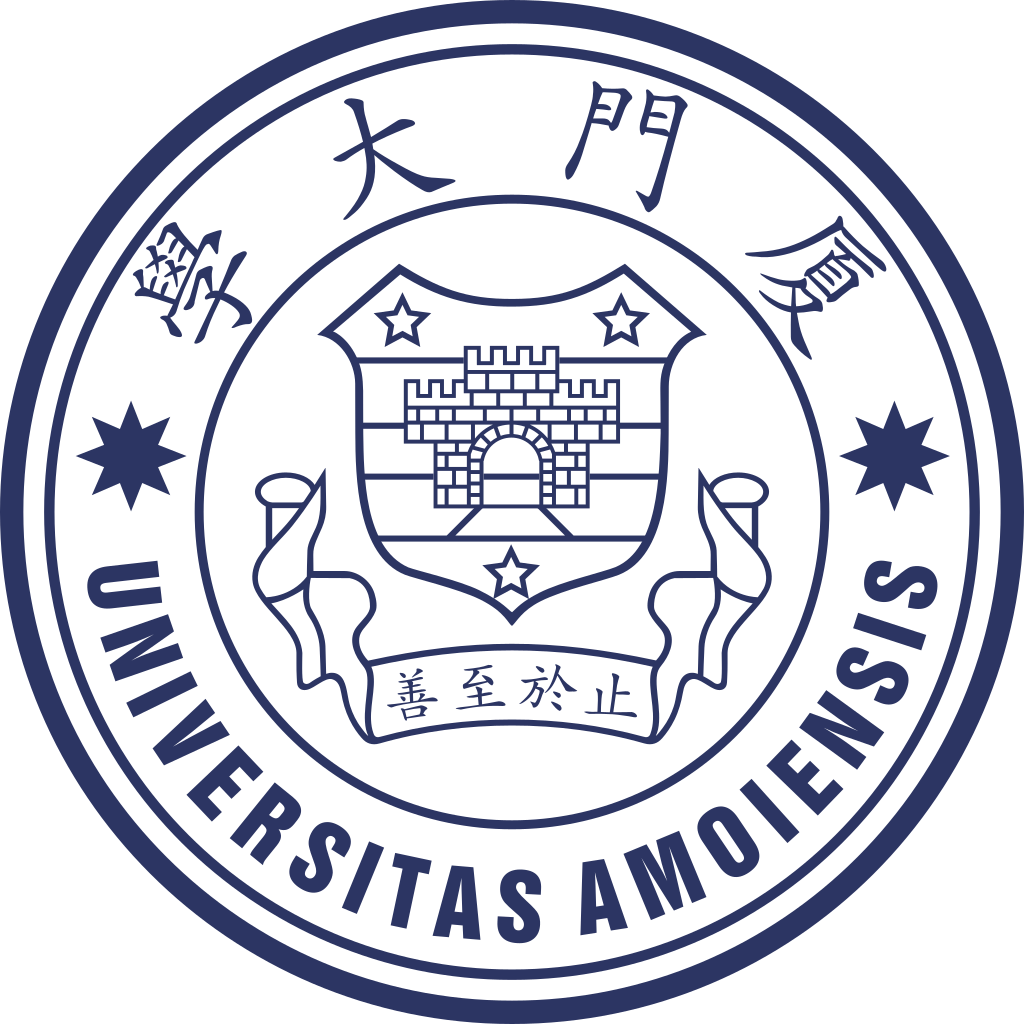}%
  \headerlogospace{2.8mm}%
  \headerlogo[-0.05\height]{12.8mm}{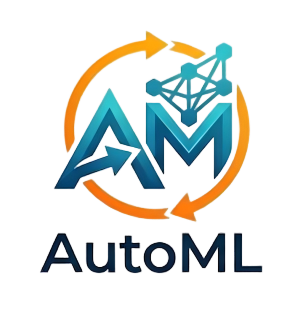}%
}
\setrightheadericon{}
\setrunningheadericon{%
  \headerlogospace{1.2mm}%
  \headerlogo[-0.03\height]{8.2mm}{assets/branding/automl.png}%
}
\setheadergroupname{MAC-AutoML}

\title{OmniScope: Modality-Decoupled Token Compression for Omnimodal Large Language Models}

\setfrontauthors{%
  \authorrow{%
    \authorentry{Jinsen Su}{1,2},\authorsep
    \authorentry{Yongdong Luo}{1,3,4},\authorsep
    \authorentry{Yuexiao Ma}{1,3},\authorsep
    \authorentry{Yibo Hu}{4}}%
  \authorrow{%
    \authorentry{Meiguang Jin}{4},\authorsep
    \authorentry{Xiawu Zheng}{1,2,\corremailmark}}%
}
\setfrontaffiliations{%
  \affiliationline{\authormark{1}Media Analytics and Computing Lab, Xiamen University, Xiamen, China}
  \affiliationline{\authormark{2}Institute of Artificial Intelligence, Xiamen University, Xiamen, China}
  \affiliationline{\authormark{3}School of Informatics, Xiamen University, Xiamen, China}
  \affiliationline{\authormark{4}Alibaba Group}
}
\setfrontcontact{%
  \contactline{\textbf{\corremailmark\ Corresponding Author}}
}
\usecustomauthorlayout

\abstract{

Existing token compression methods for omnimodal large language models typically rely on one modality to determine what to retain in the other. We show that this assumption often breaks down: for the same query, audio and video relevance often peaks at different moments. This cross-modal salience mismatch makes unidirectional guidance prone to discarding answer-critical cues under aggressive compression. We propose OmniScope, a training-free token compression framework that uses the query as a shared semantic anchor while estimating relevance separately for audio and video. OmniScope allocates modality-specific token budgets, prunes visual tokens with an anchor-delta strategy that preserves both global context and temporal changes, and merges audio tokens within each second to reduce redundancy while maintaining temporal continuity. Across four audio-video benchmarks and two Qwen2.5-Omni model scales, OmniScope achieves the best average accuracy across all compression settings. At 25\% overall token retention, it delivers up to 3.53x prefill speedup and more than 15\% GPU memory reduction, with only a 0.35-point drop in average accuracy. These results suggest a simple design principle for OmniLLM inference: share the query across modalities, but not the salience estimates. The code is available at the GitHub repository: \url{https://github.com/MAC-AutoML/OmniScope}.

}

\begin{document}

\maketitle

\begin{figure}[t]
\centering
\includegraphics[width=0.88\textwidth]{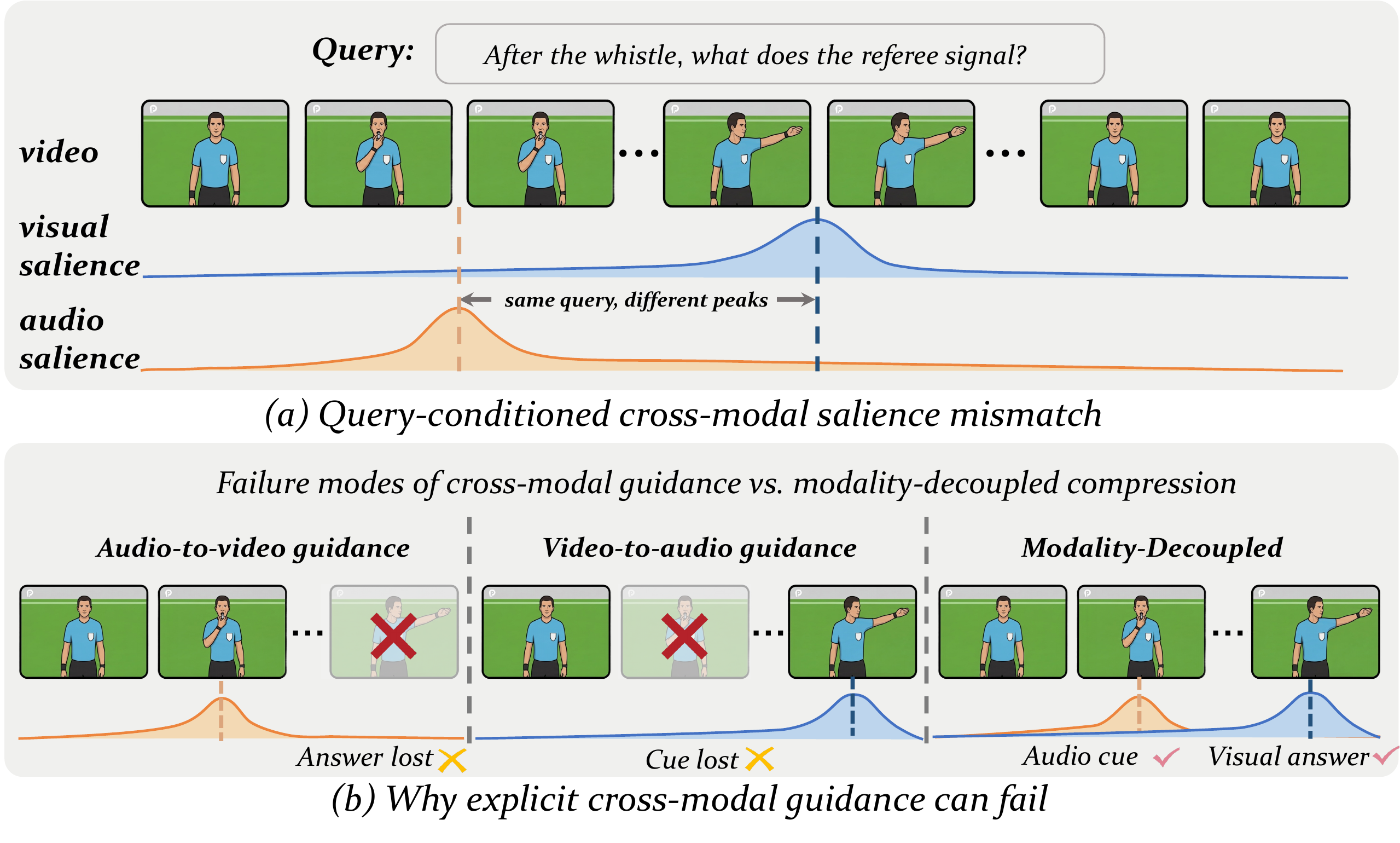}
\caption{One query, two modalities, and often two very different sparsity patterns. (a) Audio salience peaks early (whistle) while visual salience peaks late (hand signal), illustrating cross-modal salience mismatch. (b) Unidirectional cross-modal guidance loses critical information, while modality-decoupled compression preserves both.}
\vspace{-1.3em}
\label{fig:salience-mismatch}
\Description{The figure has two panels. Panel (a) shows audio and visual salience score curves for a sample video. The audio salience curve peaks in the first half of the timeline, corresponding to a whistle sound, while the visual salience curve peaks in the second half, corresponding to a hand signal gesture; the two peaks have minimal overlap, illustrating cross-modal salience mismatch. Panel (b) compares three compression strategies: unidirectional audio-guided compression preserves audio-critical information but discards visually critical information; unidirectional visual-guided compression preserves visually critical information but loses audio-critical information; modality-decoupled compression independently preserves the critical information of both modalities.}
\end{figure}
\section{Introduction}
\label{sec:introduction}

Omnimodal large language models (OmniLLMs)~\cite{xu2025qwen2,fu2025vita,sun2024video,xie2024mini,zhao2025r1,ye2025omnivinci,liu2025javisgpt,xu2025qwen3,tang2025video,yang2025humanomniv2} can jointly process visual, auditory, and textual modalities, enabling end-to-end audio-video understanding and representing a major advance in multimodal intelligence.
Compared with models that handle only images or text, OmniLLMs capture the synergistic semantics between visual scenes and acoustic signals, delivering significant advantages in applications such as video question answering, content moderation, and real-time interaction.

However, deploying OmniLLMs in practice poses severe efficiency challenges. A typical OmniLLM decomposes a video into visual frames and an audio stream, and encodes them into token sequences via separate encoders. As video duration increases, the total number of audio-visual tokens grows rapidly, incurring substantial computational and memory overhead while severely degrading the quality of contextual interaction, limiting real-time deployment on resource-constrained devices.

Token compression offers a direct means to alleviate this bottleneck.
In vision-only multimodal LLMs, a variety of token compression methods have been proposed to reduce inference cost by pruning or merging redundant visual tokens~\cite{chen2024image,yang2025visionzip,shang2025llava,xing2024pyramiddrop,tao2025dycoke}.
However, these methods target purely visual settings and cannot directly address the joint compression of audio and video tokens.
The few methods designed for omnimodal scenarios, such as OmniZip~\cite{tao2025omnizip}, adopt unidirectional cross-modal guidance strategies, letting one modality determine what is retained in the other.
However, in OmniLLMs, the query is shared across modalities, but the relevance pattern is often not---visually decisive moments may be acoustically unremarkable, while audio-critical segments may exhibit little visual change.
As illustrated in Fig.~\ref{fig:salience-mismatch}, this misalignment of cross-modal salience peaks means that low-salience regions in one modality may carry critical cues for the other, making unidirectional guidance prone to erroneously discarding key information under high compression ratios.

Our analysis of the attention distribution in OmniLLMs further reveals that the consequences of this mismatch are more severe than they may appear: as shown in Fig.~\ref{fig:attention}, token compression significantly strengthens cross-window modal interaction, but when unidirectional guidance discards critical tokens under salience mismatch, the enhanced cross-window attention is instead steered toward secondary cues. Conversely, if each modality retains only its own query-relevant tokens, the amplified cross-window attention naturally falls on task-relevant cues from both sides. This motivates the core design principle of OmniScope: using the query as a shared semantic anchor while letting each modality independently decide its own compression strategy, ensuring that critical information from both modalities is preserved and that the interaction enhancement brought by compression is directed toward task-relevant cues. Building on this principle, we propose OmniScope, a training-free token compression framework for OmniLLMs.

OmniScope uses the query as a shared semantic anchor while allowing audio and visual tokens to independently assess their importance and allocate compression budgets.
On this basis, the visual side employs Anchor-Delta Spatio-Temporal Compression (AD-STC) to balance global semantic coverage and temporal increments, while the audio side adopts per-second Token Merging to reduce redundancy while maintaining temporal continuity.

We validate OmniScope on four multimodal video understanding benchmarks at two model scales (7B and 3B).
OmniScope achieves the best average performance at all compression ratios, with nearly lossless accuracy at 45\% retention.
At the more aggressive 25\% retention, OmniScope still maintains significantly stronger robustness than all competing methods, while delivering up to $3.53\times$ prefilling speedup and over 15\% GPU memory reduction.

Our contributions are summarized as follows:
\begin{itemize}
    \item We identify that the unidirectional cross-modal guidance assumed by prior omnimodal compression methods is unreliable under cross-modal salience mismatch, and propose OmniScope, a training-free, modality-decoupled token compression framework that uses the query as a shared semantic anchor while allowing each modality to independently compress its tokens without explicit cross-modal guidance.
    \item We propose AD-STC, an anchor-delta spatio-temporal compression strategy that addresses the loss of global semantic coverage in existing video token compression methods.
    \item On the audio side, we introduce per-second bipartite soft matching on post-encoder temporal embeddings, addressing the coarse granularity of existing audio compression while preserving temporal continuity.
    \item On four benchmarks and two model scales, OmniScope achieves the best average performance at all compression ratios, with nearly lossless accuracy at 45\% retention.
\end{itemize}

\section{Related Work}

\subsection{Omnimodal Large Language Models}
Omnimodal large language models (OmniLLMs)~\cite{xu2025qwen2,fu2025vita,sun2024video,xie2024mini,zhao2025r1,ye2025omnivinci,liu2025javisgpt,xu2025qwen3,tang2025video,yang2025humanomniv2} aim to unify visual, auditory, and textual modalities within a single model for end-to-end audio-video understanding. Compared with vision-language models~\cite{li2024llava,chen2024far,liu2023visual,lin2024video,dai2023instructblip,wang2024qwen2} that handle only visual and textual modalities, OmniLLMs capture the synergistic semantics between visual scenes and acoustic signals, delivering significant advantages in applications such as video question answering, content moderation, and real-time interaction. Representative open-source systems include Qwen2.5-Omni~\cite{xu2025qwen2} and VITA-1.5~\cite{fu2025vita}, while proprietary systems such as GPT-4o~\cite{hurst2024gpt} and Gemini~\cite{team2023gemini} further demonstrate the effectiveness of this paradigm. These systems typically employ modality-specific encoders and a time-window interleaving mechanism to organize multimodal tokens. As video duration increases, the total number of audio-visual tokens grows rapidly, 
incurring substantial computational and memory overhead. Existing approaches 
improve efficiency through multiple techniques, including token 
compression~\cite{bolya2022token,tao2025dycoke,shao2025holitom}, model 
quantization~\cite{ma2024affinequant,ma2023ompq,xiao2023smoothquant}, 
and network pruning~\cite{zheng2021information,ma2023llm,frantar2023sparsegpt}. Among them, token 
compression directly addresses the growing audio-visual sequence length, 
making it a key problem~\cite{jin2025efficient} for efficient OmniLLM inference.

\subsection{Token Compression in Multimodal Models}
In the visual domain, token compression methods have been extensively studied~\cite{shao2025tokens} and mainly include similarity-based token merging~\cite{bolya2022token,shao2025holitom}, query-based text-guided compression~\cite{luo2026quota,zhang2024sparsevlm,song2024moresimpleeffectivetoken,sun2025lvpruning}, and attention-based token pruning~\cite{chen2024image,yang2025visionzip,shang2025llava,xing2024pyramiddrop}. In the video domain, spatio-temporal compression methods~\cite{tao2025dycoke,shao2025holitom,weng2024longvlm,huang2025prunevid} further exploit inter-frame redundancy to reduce the token count. However, existing video spatio-temporal compression methods typically apply the same compression strategy to all temporal incremental frames, overlooking the differences in semantic density and temporal variation across frames, which tends to cause a loss of global semantic coverage under high compression ratios.
On the audio side, representative token reduction techniques include adjacent token merging~\cite{li2023accelerating} and importance-based pruning~\cite{lin2025speechprune,lee2025token}, which are constrained by the encoder's internal structure or coarse-grained compression strategies, making it difficult to flexibly adapt to the compression ratios required by downstream LLMs and to minimize information loss during compression.
Recent work has begun to explore audio--visual joint compression. OmniZip~\cite{tao2025omnizip} is a training-free method that uses audio importance scores to guide video token pruning, while OmniSIFT~\cite{ding2026omnisift} requires end-to-end fine-tuning of both an additional compression module and the LLM decoder, and employs vision-guided modality-asymmetric compression. Although the two methods guide compression in opposite directions, both rely on unidirectional cross-modal guidance---using one modality's importance to determine the retention strategy of the other. Moreover, the training dependency of OmniSIFT limits its generalizability across different models.

\section{Method}

\subsection{Preliminaries}
\label{sec:preliminaries}

Omnimodal large language models (OmniLLMs) aim to jointly process visual, auditory, and textual modalities for end-to-end audio-video understanding. Taking Qwen2.5-Omni~\cite{xu2025qwen2} as a representative example, a typical OmniLLM comprises four components: a vision encoder $\Phi_v$, an audio encoder $\Phi_a$, a projector, and an LLM backbone. Given a video, the system first decomposes it into a sequence of video frames $X_{\text{vid}}$ and the corresponding audio segments $X_{\text{aud}}$. The vision encoder and audio encoder convert them into token sequences respectively:
\begin{equation}
    Z_v = \Phi_v(X_{\text{vid}}) \in \mathbb{R}^{N_v \times D}, \quad Z_a = \Phi_a(X_{\text{aud}}) \in \mathbb{R}^{N_a \times D},
\end{equation}
where $N_v$ and $N_a$ denote the number of visual and audio tokens, and $D$ is the embedding dimension. The projector maps both token types into the LLM's embedding space, where they are processed together with text tokens to generate a response.

To organize the audio-video tokens, OmniLLMs adopt a \textit{time-window interleaving} mechanism. Specifically, the audio and video streams are segmented into $W$ time windows of fixed duration. Within each window, co-temporal visual tokens and audio tokens are aligned and concatenated into a cross-modal block; these blocks are then arranged chronologically to form the full input sequence to the LLM. Let the $i$-th time window contain $n_v^{(i)}$ visual tokens and $n_a^{(i)}$ audio tokens. The sequence received by the LLM can be expressed as:
\begin{equation}
    X_{\text{LLM}} = \big[B_1, B_2, \ldots, B_W\big], \quad B_i = \big[Z_v^{(i)};\, Z_a^{(i)}\big],
\end{equation}
where $[\,\cdot\,;\,\cdot\,]$ denotes concatenation along the sequence dimension. Under this mechanism, visual and audio tokens within each window jointly participate in the LLM's self-attention computation. As the video duration increases, the total number of audio-visual tokens grows rapidly, not only incurring substantial computational and memory overhead but also allowing a large volume of redundant tokens to participate in attention computation, posing challenges to both the efficiency and quality of the model's reasoning.

\begin{figure}[!t]
\centering
\includegraphics[width=0.8\textwidth]{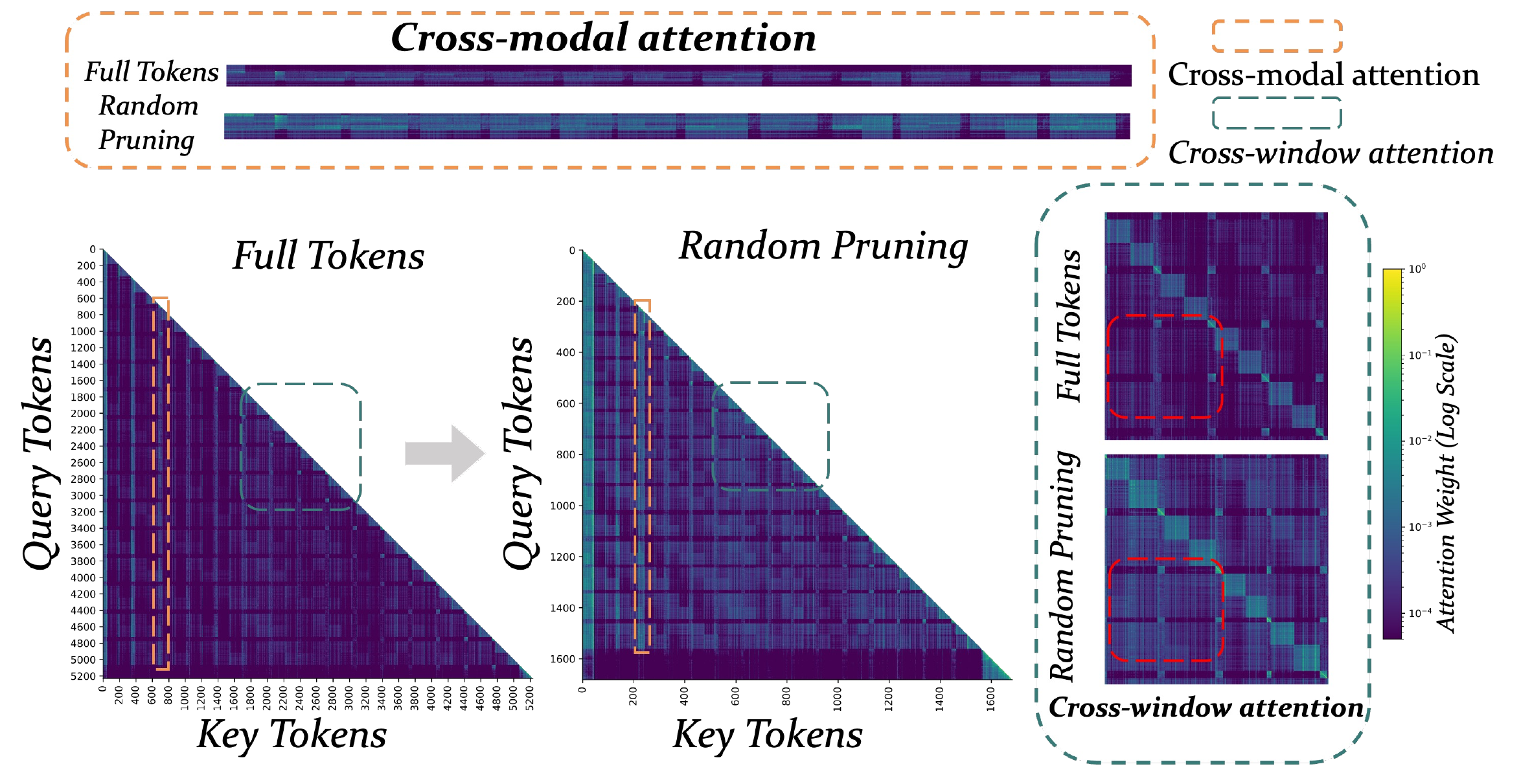}
\caption{Comparison of attention maps at Layer 3 of Qwen2.5-Omni-3B. Under the Full Token setting, attention is concentrated locally; after random pruning, both cross-window (the off-diagonal regions highlighted by red dashed boxes are notably enhanced) and cross-modal attention are significantly strengthened.}
\vspace{-1.3em}
\label{fig:attention}
\Description{Side-by-side comparison of two attention heatmaps, both from Layer 3 of Qwen2.5-Omni-3B. The left heatmap shows the Full Token (uncompressed) setting, where attention weights are highly concentrated along the diagonal within local temporal windows, and cross-window and cross-modal attention are extremely weak. The right heatmap shows the setting after random token pruning, where notable attention activation appears in off-diagonal cross-window regions (highlighted by red dashed boxes) and cross-modal attention is also significantly strengthened, indicating that compression reduces the dilution of the attention budget by redundant tokens within each window.}
\end{figure}

\subsection{Motivation: From Attention Dilution to Salience Mismatch}
\label{sec:motivation}

Since tokens within each temporal window jointly participate in softmax-normalized self-attention, redundant tokens within the window can dilute the attention budget available to truly important cross-window tokens. To verify this, we visualize the attention distribution across different layers of Qwen2.5-Omni-3B, as illustrated in Fig.~\ref{fig:attention} (showing Layer~3 as a representative example). Under the uncompressed Full Token setting, attention is heavily concentrated within the diagonal blocks of local temporal windows, while both cross-window attention and cross-modal attention are extremely weak. After random token compression, the number of tokens within each window is reduced, and notable attention activation emerges in the off-diagonal cross-window regions, with cross-modal attention similarly strengthened. This finding demonstrates that compression can effectively enhance cross-window modal interaction. Unidirectional cross-modal guidance methods~\cite{tao2025omnizip} essentially exploit this property: by using one modality as the guide and removing redundant tokens of the other modality around it, these methods strengthen the interaction between the guiding modality's information and other temporal windows, thereby maintaining accuracy after compression. However, since the two modalities often exhibit very different relevance patterns for the same query (as illustrated in Fig.~\ref{fig:salience-mismatch}), the importance scores of the guiding modality cannot reliably reflect the information value of the other modality. We further quantify the pervasiveness of this mismatch: across 1{,}197 query-video pairs, approximately 78.3\% exhibit only weak correlation between visual and audio salience scores (see Appendix~F). Under high compression ratios, this mismatch causes critical tokens in the non-guiding modality to be erroneously discarded, resulting in the loss of key cross-modal information. This directly inspires the modality-decoupled design of OmniScope: using the query as a shared semantic anchor, each modality independently removes task-irrelevant redundant tokens, so that the retained core tokens cluster around the same task semantics.

\subsection{OmniScope}
\label{sec:OmniScope}

\begin{figure*}[t]
    \centering
    \includegraphics[width=0.96\textwidth]{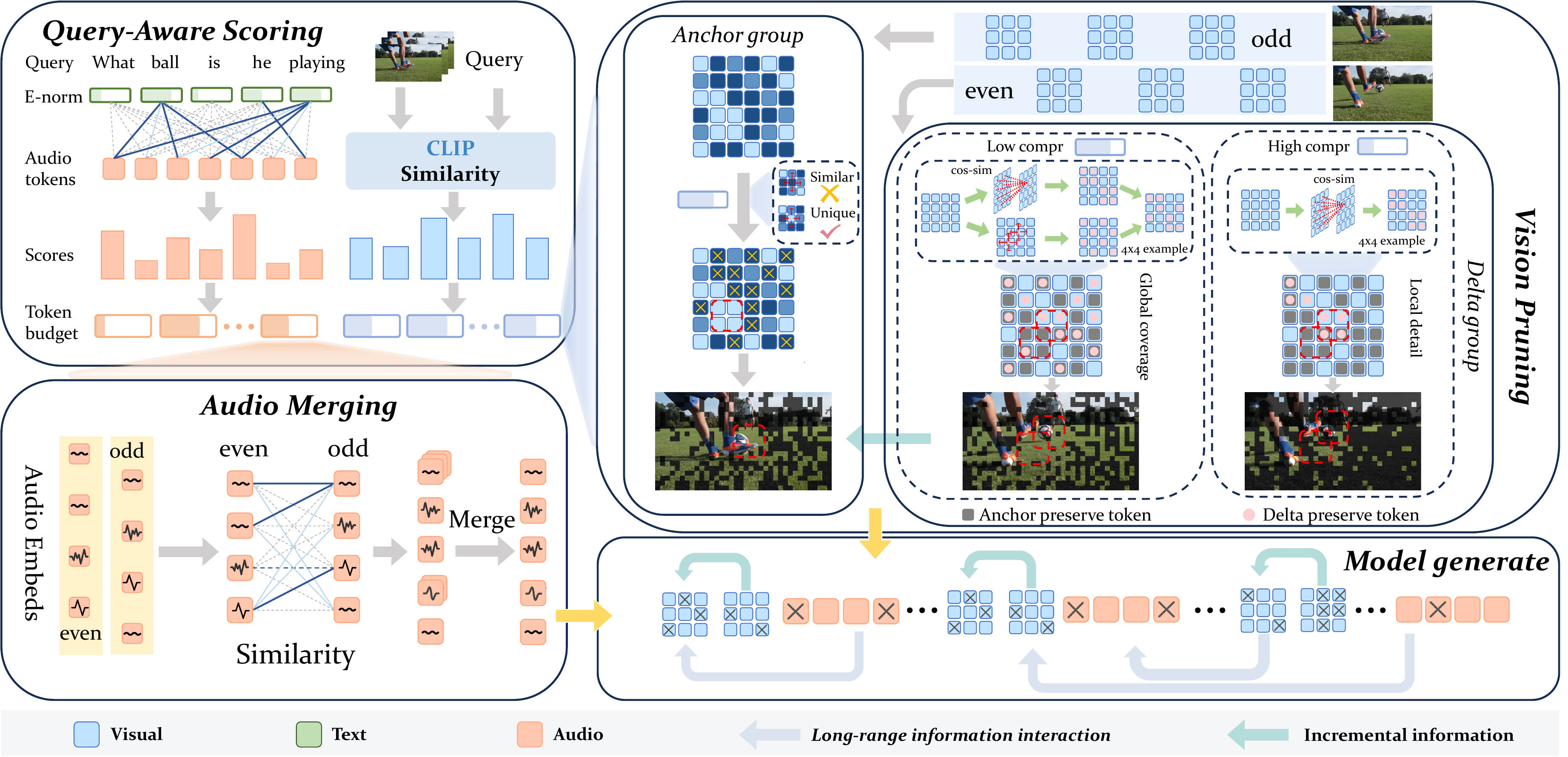}
    \caption{Overview of OmniScope. OmniScope operates in three stages: (1)~Query-Aware Scoring independently estimates the importance of visual and audio tokens relative to the query and allocates per-position compression budgets; (2)~Vision Pruning balances global semantic coverage and temporal increments through anchor--delta frame alternation; (3)~Audio Merging fuses redundant tokens via per-second bipartite soft matching.}
    
    \label{fig:pipeline}
    \Description{Overall pipeline of OmniScope, organized from left to right in three stages. Stage 1, Query-Aware Scoring: the upper branch uses CLIP to compute cosine similarity between video frames and the query text, producing per-frame visual importance scores; the lower branch uses the OmniLLM's own audio encoder embeddings and query embeddings to compute cross-modal similarity, producing per-second audio importance scores. The two sets of scores are generated independently and each proportionally allocates token retention budgets to its respective temporal positions. Stage 2, Vision Pruning: an Anchor-Delta alternating strategy is applied---even-indexed frames serve as Anchor frames, retaining the most spatially information-dense tokens via density-peak clustering; odd-indexed frames serve as Delta frames, adaptively switching scoring strategies according to the compression ratio to balance capturing temporal increments relative to the preceding Anchor frame and maintaining global semantic coverage. Stage 3, Audio Merging: within each one-second audio segment, tokens are partitioned into a source set and a destination set by alternating index, bipartite soft matching is performed, and the most similar token pairs are merged, reducing the token count while preserving temporal continuity.}
\end{figure*}

We introduce OmniScope, a training-free, query-aware omni-token compression framework. As illustrated in Fig.~\ref{fig:pipeline}, OmniScope operates in three stages: (1)~Query-Aware Scoring, which independently assesses the importance of visual and audio tokens based on their relevance to the input query and allocates per-position compression budgets accordingly; (2)~Vision Pruning, which compresses visual tokens through anchor--delta interleaved spatio-temporal pruning; and (3)~Audio Merging, which compresses audio tokens through per-second bipartite soft matching. 
\subsubsection{Visual Scoring.}
To assess the semantic relevance of each video frame to the input query, we adopt CLIP~\cite{radford2021learning} as the visual scorer (Fig.~\ref{fig:pipeline}, Query-Aware Scoring, right). This scorer operates independently of the OmniLLM's vision encoder, thereby decoupling importance estimation from model inference. Given sampled video frames and a query $q$, we extract the image feature $\mathbf{f}_I^{(t)} \in \mathbb{R}^{D_c}$ for the $t$-th frame and the query feature $\mathbf{f}_Q \in \mathbb{R}^{D_c}$, and compute a frame-level importance score via cosine similarity:
\begin{equation}
    s_v^{(t)} = \frac{\mathbf{f}_I^{(t)} \cdot \mathbf{f}_Q}{\|\mathbf{f}_I^{(t)}\|_2 \; \|\mathbf{f}_Q\|_2}.
\end{equation}

\subsubsection{Audio Scoring.}
A natural counterpart on the audio side would be an external cross-modal model such as CLAP~\cite{wu2023large}. However, models of this family are designed for clip-level audio--text alignment and typically operate at a granularity of several seconds or longer, whereas OmniScope requires per-second or even per-token importance estimation. This coarse temporal resolution makes CLAP ill-suited for the fine-grained scoring needed by our compression framework.

We instead observe that the OmniLLM's audio encoder outputs, after passing through the projector, are already mapped into the same embedding space as the LLM tokens, allowing the relevance between audio tokens and the query text to be computed directly without an external model (Fig.~\ref{fig:pipeline}, Query-Aware Scoring). In principle, the same strategy could be applied to the visual side, but as shown in Appendix~D, the model's internal vision--text alignment is not yet sufficient for reliable scoring; the visual side therefore still relies on the external CLIP scorer.

Concretely, let the audio encoder output be $\mathbf{Z}_a \in \mathbb{R}^{N_a \times D}$ and the query embedding obtained from the LLM's embedding layer be $\mathbf{E}_Q \in \mathbb{R}^{L \times D}$. Prior work has shown that the squared norm of word embeddings encodes the information gain conveyed by the word, with semantically informative words exhibiting larger norms and high-frequency function words exhibiting smaller norms~\cite{oyama2023norm}. We therefore use the embedding norm as a simple gating signal to down-weight uninformative tokens:
\begin{equation}
    g_j = \frac{\|\mathbf{E}_Q^{(j)}\|}{\max_k \|\mathbf{E}_Q^{(k)}\|}.
\end{equation}

We compute the gated cross-modal similarity matrix and aggregate it into a per-token importance score:
\begin{equation}
    S_{ij} = \bar{\mathbf{z}}_a^{(i)} \cdot \bar{\mathbf{e}}_Q^{(j)},
    w_{ij} = \mathrm{softmax}_{j}\!\left(\frac{g_j \, S_{ij}}{\tau}\right), 
    s_{\text{audio}}^{(i)} = \textstyle\sum\nolimits_{j}\, w_{ij} \, S_{ij},
\end{equation}
where $\bar{\cdot}$ denotes $\ell_2$-normalization, and $i$ indexes audio tokens while $j$ indexes query tokens. Scores are then aggregated at one-second granularity by averaging the top-$k$ token scores within each second, yielding per-second importance values $s_a^{(s)}$. Since audio signals exhibit strong temporal continuity, the context surrounding a high-scoring second is often equally informative. We therefore apply a decaying score propagation to positions near key seconds.
We initialize $\hat{s}_a^{(s)} \leftarrow s_a^{(s)}$ for all $s$,
and let $\mathcal{H}$ denote the set of seconds whose score exceeds the
$P$-th percentile. For each second $s$, we compute a boost factor from
all nearby high-scoring seconds and apply it:
\begin{equation}
    b_s = \max_{s' \in \mathcal{H},\; 0 < |s-s'| \leq R}
          \bigl(1 + (\beta - 1) \cdot \gamma^{\,|s-s'|}\bigr), \quad
    \hat{s}_a^{(s)} \leftarrow \hat{s}_a^{(s)} \cdot b_s,
\end{equation}
where $\beta$ controls the base boost magnitude, $\gamma$ is
the distance decay rate, and $R$ is the propagation radius.
\subsubsection{Token Budget Allocation.}
Once the importance scores are obtained, each modality is allocated a compression budget independently. Given a modality's global retention ratio $1{-}\rho$, the total number of retained tokens is $N^{\mathrm{budget}} = (1{-}\rho) \cdot N_{\mathrm{total}}$. Based on the position-level importance scores $\{s^{(i)}\}$ (frame-level for vision, second-level for audio), the number of tokens retained at each position is allocated proportionally:
\begin{equation}
    n_{\mathrm{keep}}^{(i)} = \left\lfloor \frac{s^{(i)}}{\sum_j s^{(j)}} \cdot N^{\mathrm{budget}} + 0.5 \right\rfloor.
\end{equation}
The compression ratios $\rho_v$ and $\rho_a$ are configured independently---each modality determines its retained token count based solely on its own query relevance, without cross-modal comparison.

\subsection{AD-STC}
\label{sec:AD-STC}

This section addresses \emph{which} tokens to retain within each frame. Existing video token compression methods~\cite{tao2025dycoke,shao2025holitom,weng2024longvlm,huang2025prunevid} typically apply a uniform strategy to temporal incremental frames, which tends to lose global semantic coverage under high compression ratios. We therefore propose AD-STC (Anchor-Delta Spatio-Temporal Compression), which alternately assigns frames to two complementary roles (Fig.~\ref{fig:pipeline}, Vision Pruning). \textit{Anchor frames} (even-indexed) retain the most spatially 
distinctive tokens, capturing the core semantics of the current moment. \textit{Delta frames} (odd-indexed) retain only the tokens that represent the most significant information increment relative to the preceding anchor frame, and adaptively switch scoring strategies according to the compression ratio---maintaining both global coverage and local detail when the budget is generous, while concentrating the limited resources on capturing dense, locally complete temporal increments to avoid information fragmentation when the budget is tight. Alternating between the two roles ensures a balanced preservation of both spatial and temporal information.

\subsubsection{Anchor Frame: Spatial Density Selection.}
Anchor frames aim to retain the most spatially informative tokens within the allocated budget. We adopt a density-peak clustering approach with $k$-nearest neighbors (DPC-KNN)~\cite{du2016study}. For each token $\mathbf{h}_i$ in the frame, we compute its local density as the mean similarity to its $k$ nearest neighbors:
\begin{equation}
    d_i = \frac{1}{k} \sum_{\mathbf{h}_j \,\in\, \mathrm{KNN}(\mathbf{h}_i)} \mathrm{sim}(\mathbf{h}_i,\, \mathbf{h}_j).
\end{equation}
A high $d_i$ indicates that the token closely resembles its neighbors and is therefore spatially redundant, while a low $d_i$ indicates that it is relatively isolated in the feature space and carries distinctive semantic content. We rank all tokens by $d_i$ in ascending order and retain the $n_{\mathrm{keep}}^{(t)}$ tokens with the lowest values.
\subsubsection{Delta Frame: Adaptive Temporal Selection.}
Delta frames aim to capture the information increment relative to the preceding anchor frame. We adaptively switch between two scoring strategies according to the retention ratio $r = n_{\mathrm{keep}}^{(t)} / n_{\mathrm{total}}$: when $r \geq \tau_r$, a joint spatio-temporal criterion is used; when $r < \tau_r$, a temporal-difference criterion is applied instead.

\paragraph{Low compression (generous budget).}\;
When the retention ratio is high, we employ a joint irreplaceability score that accounts for both spatial and temporal dimensions simultaneously. Let $\{\mathbf{h}_i\}$ denote the tokens of the current delta frame and $\{\mathbf{h}_k^{\mathrm{prev}}\}$ denote the tokens of the preceding anchor frame. For each token $\mathbf{h}_i$, we compute a spatial replaceability term and a temporal replaceability term:
\begin{equation}
    \alpha_i^{\mathrm{intra}} = \max_{j \neq i} \operatorname{sim}(\mathbf{h}_i, \mathbf{h}_j), \quad \alpha_i^{\mathrm{inter}} = \max_{l} \operatorname{sim}(\mathbf{h}_i, \mathbf{h}_l^{\mathrm{prev}}),
\end{equation}
where $\alpha_i^{\mathrm{intra}}$ measures how well a token can be substituted by another token in the same frame (a high value indicates spatial redundancy), and $\alpha_i^{\mathrm{inter}}$ measures how well it can be substituted by a token in the preceding anchor frame (a high value indicates temporal redundancy).The two terms are fused multiplicatively into an overall replaceability score:
\begin{equation}
    \mathrm{replaceability}_i = \alpha_i^{\mathrm{intra}} \times \alpha_i^{\mathrm{inter}}.
\end{equation}
The product attains a high value only when a token is highly replaceable along \emph{both} dimensions simultaneously; it is then deemed redundant and removed. A token that is replaceable along only one dimension---for instance, spatially similar to its neighbors yet temporally novel---still yields a low product and is therefore retained. Consequently, removed tokens must satisfy both spatial and temporal redundancy. Compared with single-dimension criteria, the multiplicative fusion yields a more dispersed retained set in the spatio-temporal domain, providing more uniform global coverage when the token budget is sufficient. Accordingly, we rank all tokens by replaceability in ascending order and retain the $n_{\mathrm{keep}}^{(t)}$ tokens with the lowest scores.

\paragraph{High compression (tight budget).}\;
When the retention ratio is low, the token budget is severely constrained. Under such conditions, the joint spatio-temporal criterion tends to scatter the few retained tokens across both spatial and temporal dimensions, yielding sparse, fragmented information that lacks local coherence. We therefore concentrate the limited budget on capturing dense, locally complete temporal increments: we compute the per-position cosine similarity between the current and the preceding frame, $c_i = \mathrm{sim}(\mathbf{h}_i,\, \mathbf{h}_i^{\mathrm{prev}})$, and retain the $n_{\mathrm{keep}}^{(t)}$ tokens with the lowest similarity---i.e., those that have changed the most.

\subsection{Audio Token Compression}
\label{sec:audio_compression}

Due to the short-term stationarity of audio signals~\cite{rabiner1993fundamentals}, temporally adjacent tokens overlap substantially yet each carries subtle acoustic differences; directly pruning tokens would cause irrecoverable information gaps along the temporal axis. Unlike the coarse compression granularity of existing audio compression methods~\cite{li2023accelerating,lin2025speechprune}, we perform per-second token merging on the one-dimensional temporal embeddings after the encoder, fusing overlapping information at a finer granularity rather than discarding it, thereby minimizing information loss and preserving temporal coverage integrity (Fig.~\ref{fig:pipeline}, Audio Merging).

Given the per-second token budgets $\{n_{\mathrm{keep}}^{(s)}\}$ produced by the allocation described in Sec.~\ref{sec:OmniScope}, we perform token merging via bipartite soft matching~\cite{bolya2022token} independently within each one-second segment. The $N$ tokens in a segment are first partitioned into two disjoint sets by alternating index: a \emph{source} set $\mathcal{A} = \{\mathbf{h}_0, \mathbf{h}_2, \ldots\}$ (even-indexed) and a \emph{destination} set $\mathcal{B} = \{\mathbf{h}_1, \mathbf{h}_3, \ldots\}$ (odd-indexed). A similarity matrix between the two sets is then computed as:
\begin{equation}
    \mathbf{M} = \bar{\mathbf{H}}_{\mathcal{A}} \;\bar{\mathbf{H}}_{\mathcal{B}}^{\top},
\end{equation}
where $\bar{\cdot}$ denotes $\ell_2$-normalization. Each source token is paired with its most similar destination token. Among all such pairs, the $m = N - n_{\mathrm{keep}}^{(s)}$ pairs with the highest similarity scores are selected for merging. Each selected source token is fused into its matched destination through averaging:
\begin{equation}
    \mathbf{h}_{\mathrm{dst}}^{\prime} = \frac{\mathbf{h}_{\mathrm{dst}} \;+\; \displaystyle\sum_{i \,\in\, \mathcal{S}(\mathrm{dst})} \mathbf{h}_i}{\;1 + \lvert\mathcal{S}(\mathrm{dst})\rvert\;},
\end{equation}
where $\mathcal{S}(\mathrm{dst})$ is the set of source tokens assigned to the destination token. If the required reduction exceeds $\lfloor N/2 \rfloor$, this step is applied iteratively.


\begin{table*}[t]
\caption{Comparison on omnimodal (audio \& video) QA benchmarks at 45\% and 25\% retained ratios. The \textbf{best} result among compression methods for each metric is bolded. $\Delta$ denotes the difference relative to the Full Tokens baseline.}
\label{tab:main}
\centering
\begin{tabular*}{\textwidth}{@{\hspace{6pt}\extracolsep{\fill}}llcccccr@{\hspace{6pt}}}
\toprule
Method & Ratio & WorldSense & DailyOmni & OmniVideoBench & Video-MME & Avg. & $\Delta$ \\
\midrule
\multicolumn{8}{c}{\emph{Qwen2.5-Omni-7B}} \\
\midrule
Full Tokens & 100\% & 46.0 & 62.0 & 34.1 & 63.3 & 51.35 & $-$ \\
\midrule
Random             & 45\% & 44.7 & 58.7 & 34.2          & 63.3          & 50.23          & $-1.12$ \\
FastV (A\&V)       & 45\% & 45.8 & 59.5 & \textbf{35.8} & 63.4          & 51.13          & $-0.22$ \\
OmniZip            & 45\% & 45.7 & 60.0 & 35.6          & 63.3          & 51.15          & $-0.20$ \\
\textbf{OmniScope (Ours)} & 45\% & \textbf{46.3} & \textbf{60.5} & 35.5 & \textbf{64.3} & \textbf{51.65} & $\mathbf{+0.30}$ \\
\midrule
Random             & 25\% & 44.9 & 56.9 & 34.9          & 62.4          & 49.78          & $-1.57$ \\
FastV (A\&V)       & 25\% & 44.7 & 58.4 & 35.6          & 63.4          & 50.53          & $-0.82$ \\
OmniZip            & 25\% & 44.8 & 57.8 & 33.7          & 62.9          & 49.80          & $-1.55$ \\
\textbf{OmniScope (Ours)} & 25\% & \textbf{45.7} & \textbf{58.8} & \textbf{35.8} & \textbf{63.7} & \textbf{51.00} & $\mathbf{-0.35}$ \\
\midrule
\multicolumn{8}{c}{\emph{Qwen2.5-Omni-3B}} \\
\midrule
Full Tokens & 100\% & 46.1 & 60.7 & 32.7 & 61.1 & 50.15 & $-$ \\
\midrule
Random             & 45\% & 44.1 & 57.6 & 32.4          & 60.9          & 48.75          & $-1.40$ \\
FastV (A\&V)       & 45\% & 44.2 & \textbf{59.7} & 33.0 & 61.2          & 49.53          & $-0.62$ \\
OmniZip            & 45\% & 45.4 & 59.1 & \textbf{33.2} & 61.1          & 49.70          & $-0.45$ \\
\textbf{OmniScope (Ours)} & 45\% & \textbf{46.1} & 59.3 & \textbf{33.2} & \textbf{62.1} & \textbf{50.18} & $\mathbf{+0.03}$ \\
\midrule
Random             & 25\% & 42.5 & 55.7 & 31.7          & 59.8          & 47.43          & $-2.72$ \\
FastV (A\&V)       & 25\% & 42.9 & 55.8 & 32.8          & \textbf{60.4} & 47.98          & $-2.17$ \\
OmniZip            & 25\% & 44.0 & 56.6 & \textbf{33.1} & 59.4          & 48.28          & $-1.87$ \\
\textbf{OmniScope (Ours)} & 25\% & \textbf{44.5} & \textbf{56.9} & \textbf{33.1} & 60.1 & \textbf{48.65} & $\mathbf{-1.50}$ \\
\bottomrule
\end{tabular*}
\end{table*}
\section{Experiments}
\label{sec:experiments}

\subsection{Experimental Setup}
\label{sec:setup}

\paragraph{Benchmarks.}
We evaluate OmniScope on four audio-video understanding benchmarks that span diverse task types and video durations:
(1)~\textbf{WorldSense}~\cite{hong2025worldsense}, which assesses joint audio-video comprehension across eight domains---Tech \& Science, Culture \& Politics, Daily Life, Film \& TV, Performance, Games, Sports, and Music;
(2)~\textbf{DailyOmni}~\cite{zhou2025daily}, which targets omnimodal question answering in everyday scenarios;
(3)~\textbf{OmniVideoBench}~\cite{li2025omnivideobench}, a comprehensive benchmark for omnimodal video understanding; and
(4)~\textbf{Video-MME}~\cite{fu2025video}, a widely adopted video understanding benchmark in which incorporating audio further improves accuracy.
Together, these benchmarks cover a broad range of video durations, task types, and audio-visual modality balance.

\paragraph{Base Models.}
Following prior work~\cite{tao2025omnizip}, we implement OmniScope on Qwen2.5-Omni~\cite{xu2025qwen2} at two parameter scales, 7B and 3B. FlashAttention-2~\cite{dao2023flashattention} is enabled in all experiments.

\paragraph{Comparison Methods.}
Token compression methods designed specifically for OmniLLMs remain scarce. We therefore adapt several representative approaches for comparison.
Full Tokens serves as the uncompressed baseline.
Random applies equal-ratio random pruning to both modalities.
OmniZip~\cite{tao2025omnizip}, the current state-of-the-art open-source training-free omnimodal token compression method, employs audio-guided dynamic video token pruning coupled
with an interleaved spatio-temporal compression module.
FastV (A\&V)~\cite{chen2024image} performs attention-based token pruning at a shallow decoder layer, applied jointly to audio and video tokens; we adapt the original implementation for FlashAttention compatibility by computing importance scores solely from the attention distribution of the last text token at the designated layer, which also reduces its memory overhead. We do not compare with OmniSIFT~\cite{ding2026omnisift}, as it requires additional training and its code is not publicly available at the time of submission, precluding reproducible inference-time comparison.

\paragraph{Implementation Details.}
Audio scoring leverages the model's own audio encoder. Visual scoring employs CLIP ViT-L/14@336px~\cite{radford2021learning}. Detailed hyperparameter settings are provided in Appendix~B. All hyperparameters are fixed across all benchmarks and both model scales without per-dataset tuning. We report results under two compression settings, referred to as 45\% retention ($\rho_v{=}0.6,\;\rho_a{=}0.25$) and 25\% retention ($\rho_v{=}0.8,\;\rho_a{=}0.35$), where the percentages denote the approximate overall token retention ratio. Video input is uniformly capped at 128 frames across all benchmarks.

\begin{figure*}[t]
  \centering
  \begin{minipage}[b]{0.44\textwidth}
    \centering
    \includegraphics[width=\linewidth]{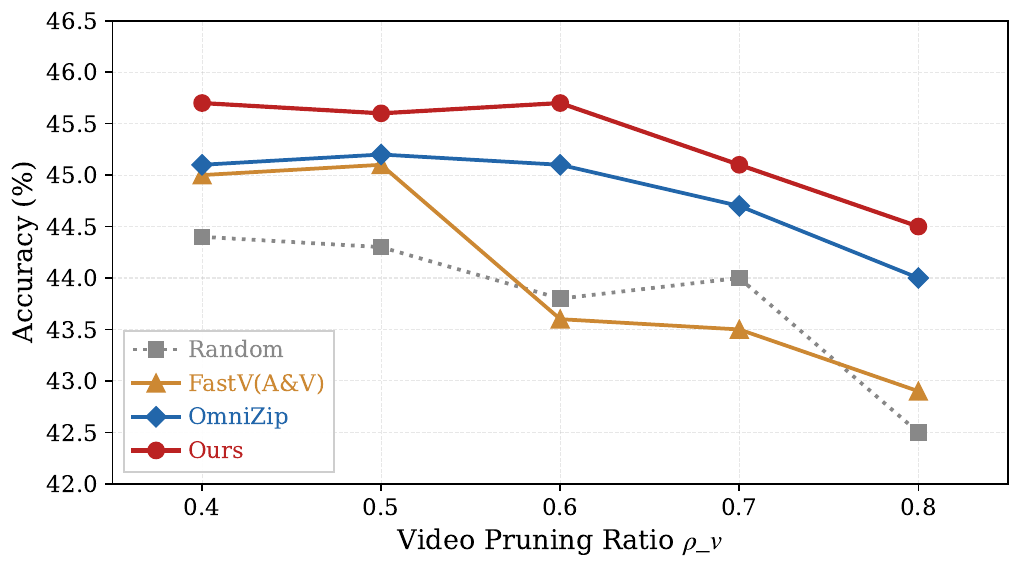}
  \end{minipage}
  \hfill
  \begin{minipage}[b]{0.44\textwidth}
    \centering
    \includegraphics[width=\linewidth]{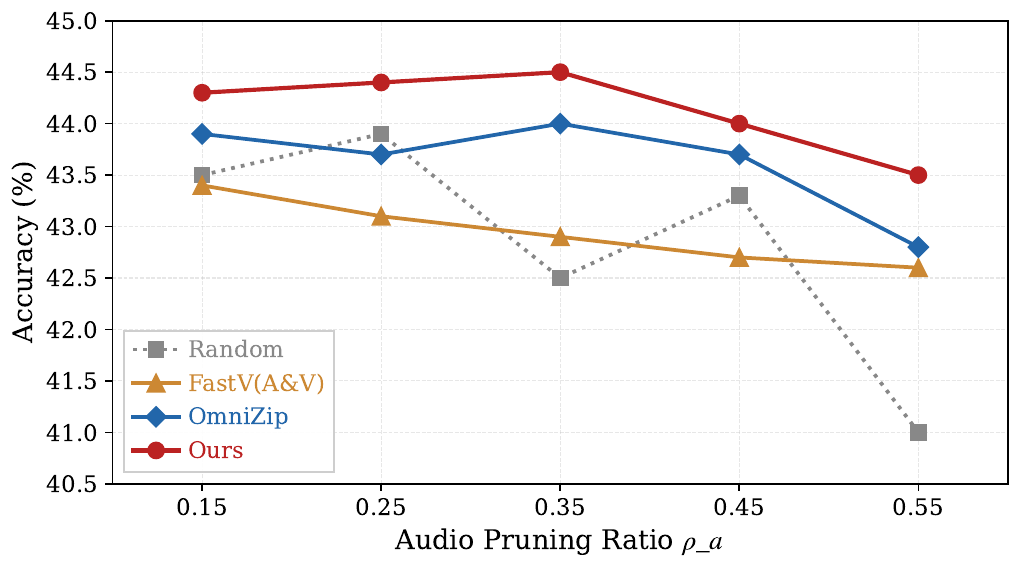}
  \end{minipage}
  \caption{\textbf{Ablation results for video and audio compression ratios}, evaluated on Qwen2.5-Omni-3B using the WorldSense benchmark. \textbf{Left:} Varying the video compression ratio $\rho_v$ with audio compression ratio $\rho_a=0.35$. \textbf{Right:} Varying the audio compression ratio $\rho_a$ with video compression ratio $\rho_v=0.8$.}
  \label{fig:ratio-ablation}
  \Description{Two line charts displayed side by side. The left chart fixes the audio compression ratio at rho_a equals 0.35 and varies the video compression ratio rho_v from 0.4 to 0.8 on the horizontal axis, with WorldSense accuracy on the vertical axis. The right chart fixes the video compression ratio at rho_v equals 0.8 and varies the audio compression ratio rho_a from 0.15 to 0.55 on the horizontal axis, with accuracy on the vertical axis. In both charts, OmniScope maintains the highest accuracy and degrades most gracefully as the compression ratio increases, while OmniZip and FastV exhibit sharp performance drops in the high-compression region.}
\end{figure*}

\subsection{Main Results}
\label{sec:main_results}
\paragraph{State-of-the-Art Compression Performance.}
As shown in Table~\ref{tab:main}, we evaluate OmniScope on four omnimodal audio-video understanding benchmarks under 45\% and 25\% token retention ratios. Across all settings, OmniScope consistently achieves the highest average accuracy among compression methods, with nearly lossless accuracy at 45\% retention.
Under the more aggressive 25\% retention setting, OmniScope incurs the smallest accuracy drop relative to the Full Tokens baseline at both model scales. For example, on the 7B model, OmniScope incurs only a 0.35-point drop while OmniZip drops by 1.55 points. This is consistent with our analysis of cross-modal salience mismatch---when the token budget is severely limited, unidirectional cross-modal guidance is more prone to erroneously discarding information at critical moments of one modality. In contrast, OmniScope allows each modality to independently assess its own query relevance, and therefore remains robust at high compression ratios.

\paragraph{Fine-Grained Category Results.}
WorldSense spans eight diverse domains. As shown in Appendix~A, OmniScope achieves the highest average accuracy at both 45\% and 25\% retention across both model scales, and remains competitive across individual domains. In contrast, competing methods exhibit uneven performance across domains---for instance, on the 7B model, OmniZip performs well on \emph{Culture \& Politics} but drops sharply on \emph{Performance} and \emph{Games} at 25\% retention. This confirms that independently compressing each modality based on query relevance provides consistent robustness across diverse audio-visual scenarios, without being sensitive to the specific modality balance of a given domain.

\paragraph{Robustness Across Compression Ratios.}
As shown in Fig.~\ref{fig:ratio-ablation}, we evaluate the stability of OmniScope by independently sweeping the video and audio compression ratios. On the visual side, as $\rho_v$ increases from 0.4 to 0.8, OmniScope consistently maintains the highest accuracy with the smallest decline. On the audio side, as $\rho_a$ increases from 0.15 to 0.55, OmniScope similarly displays the most graceful degradation curve. In contrast, OmniZip with cross-modal guidance and attention-based FastV both exhibit sharp performance drops at high compression ratios. This stability stems from the decoupled compression mechanism, which concentrates the limited token budget on the most query-relevant temporal positions, thereby preserving the most critical information within fewer tokens. This further validates that the modality-decoupled compression strategy remains robust across varying compression ratios.

\begin{table*}[t]
\renewcommand{\arraystretch}{1.25}
\centering
\begin{minipage}[t]{0.48\textwidth}
\caption{Efficiency comparison on WorldSense. GPU Mem denotes peak GPU memory usage, Prefilling Time is the time for the prefilling stage, and E2E Lat.\ is the end-to-end latency.}
\label{tab:efficiency}
\centering
\scriptsize
\resizebox{\linewidth}{!}{%
\begin{tabular}{llcccc}
\toprule
Method & Ratio & \shortstack{GPU\\Mem (G)} & \shortstack{Prefill\\(ms)} & \shortstack{E2E\\Lat.\ (s)} & Acc \\
\midrule
\multicolumn{6}{c}{\emph{Qwen2.5-Omni-7B}} \\
\midrule
Full Tokens & 100\% & 28.31 & 6299.90 & 10.965 & 46.0 \\
FastV (A\&V) & 45\% & 26.73 & 2242.46 & 4.584 & 45.8 \\
OmniZip & 45\% & 26.65 & 2179.73 & \textbf{4.508} & 45.7 \\
\textbf{OmniScope (Ours)} & 45\% & 25.14 & 2195.90 & 5.284 & \textbf{46.3} \\
\textbf{OmniScope (Ours)} & 25\% & \textbf{24.00} & \textbf{1784.45} & 4.792 & 45.7 \\
\midrule
\multicolumn{6}{c}{\emph{Qwen2.5-Omni-3B}} \\
\midrule
Full Tokens & 100\% & 18.04 & 2401.11 & 4.521 & 46.1 \\
FastV (A\&V) & 45\% & 16.82 & 1814.27 & 4.136 & 44.2 \\
OmniZip & 45\% & 16.78 & 1808.51 & \textbf{4.016} & 45.4 \\
\textbf{OmniScope (Ours)} & 45\% & 15.20 & 1845.17 & 4.728 & \textbf{46.1} \\
\textbf{OmniScope (Ours)} & 25\% & \textbf{14.24} & \textbf{1544.31} & 4.410 & 44.5 \\
\bottomrule
\end{tabular}
}
\end{minipage}\hfill
\begin{minipage}[t]{0.48\textwidth}
\caption{Ablation on query-guided scoring. WS and DO denote WorldSense and DailyOmni, respectively (same for Tables~\ref{tab:ablation-audio} and~\ref{tab:ablation-vision}). ``Uniform'' denotes equal compression across all positions without query-aware budget allocation. ``X-guide Y'' compresses Y less at positions where X is less compressed. ``X-guide Y (inv.)'' compresses Y more at positions where X is less compressed. All variants use Qwen2.5-Omni-3B with ($\rho_v{=}0.6,\;\rho_a{=}0.25$).}
\label{tab:ablation-query}
\centering
\scriptsize
\begin{tabularx}{\linewidth}{Xcccc}
\toprule
Variant & WS & DO & Avg. & $\Delta$ \\
\midrule
\textbf{Ours (full)} & \textbf{46.1} & \textbf{59.3} & \textbf{52.70} & -- \\
\midrule
A\&V uniform & 44.5 & 57.6 & 51.05 & $-1.65$ \\
Audio uniform & 45.7 & 59.0 & 52.35 & $-0.35$ \\
Visual uniform & 44.7 & 57.6 & 51.15 & $-1.55$ \\
Audio-guide Visual (inv.) & 45.2 & 58.2 & 51.70 & $-1.00$ \\
Visual-guide Audio (inv.) & 45.6 & 57.8 & 51.70 & $-1.00$ \\
Audio-guide Visual & 45.0 & 58.2 & 51.60 & $-1.10$ \\
Visual-guide Audio & 45.9 & 58.9 & 52.40 & $-0.30$ \\
\bottomrule
\end{tabularx}
\end{minipage}
\end{table*}

\subsection{Efficiency Analysis}
\label{sec:efficiency}

In the QA tasks evaluated in this work, the model generates only a short option answer, making the decoding-stage KV cache and computation overhead negligible. The speedup from token compression therefore manifests primarily in the prefilling stage. As shown in Table~\ref{tab:efficiency}, OmniScope achieves the lowest GPU memory footprint at both model scales. On the 7B model, OmniScope achieves a $2.87\times$ prefilling speedup at 45\% retention (6299\,ms $\rightarrow$ 2195\,ms), further increasing to $3.53\times$ at 25\% retention (6299\,ms $\rightarrow$ 1784\,ms), while maintaining competitive accuracy. It is worth noting that OmniZip, the current state-of-the-art omnimodal compression method, requires extracting the full attention matrix from the audio encoder to assess token importance, incurring memory overhead that grows quadratically with audio length. In contrast, OmniScope's audio scoring reuses embeddings already produced during normal inference, introducing no additional memory overhead, giving it a significant scalability advantage in long-video scenarios.

The end-to-end latency of OmniScope at 45\% retention is higher than FastV and OmniZip due to the one-time overhead of CLIP visual scoring, which runs outside the LLM inference pipeline. Nevertheless, compared with the uncompressed Full Token baseline, OmniScope still achieves approximately $2.1\times$ end-to-end speedup on the 7B model (10.965\,s $\rightarrow$ 5.284\,s). Moreover, this cost is fixed: as shown in Appendix~C, the proportion of CLIP scoring in end-to-end latency drops from approximately 13\% when generating 1 token to around 7\% at 100 tokens, decreasing steadily as the generation length grows. In longer generation scenarios such as video captioning, the prefilling speedup and KV cache savings from token compression become the dominant factors, while the fixed CLIP cost is progressively amortized, making OmniScope's end-to-end latency advantage increasingly pronounced. 

\subsection{Ablation Study}


\paragraph{Ablation Study of Query-Guided Scoring.}
Table~\ref{tab:ablation-query} compares the proposed query-aware scoring with various alternative strategies. Experimental results indicate that applying uniform compression to both modalities simultaneously at the same compression ratio causes a substantial performance degradation, with an average drop of 1.65 points. Applying uniform compression to only a single modality similarly impairs performance, confirming that dynamic compression is indispensable for both modalities. Furthermore, OmniScope consistently outperforms both vision-guided audio compression and audio-guided visual compression variants. This further validates that the scores from a single modality cannot reliably predict the token importance of another modality. In contrast, decoupling and independently compressing different modalities based on query relevance more robustly preserves key information.
\begin{table*}[t]
    \centering
    \begin{minipage}[t]{0.48\textwidth}
    \caption{Ablation on vision pruning strategy. All variants use Qwen2.5-Omni-3B with ($\rho_v{=}0.6,\;\rho_a{=}0.25$).}
    \label{tab:ablation-vision}
    \centering
    \scriptsize
    \renewcommand{\arraystretch}{1.65}
    \begin{tabularx}{\linewidth}{Xcccc}
    \toprule
    Variant & WS & DO & Avg. & $\Delta$ \\
    \midrule
    \textbf{Ours (full)} & \textbf{46.1} & \textbf{59.3} & \textbf{52.70} & -- \\
    \midrule
    Delta: temporal diff only  & 45.6 & 58.7 & 52.15 & $-0.55$ \\
    Delta: irreplaceability only & 46.0 & 57.9 & 51.95 & $-0.75$ \\
    All Anchor & 45.2 & 58.0 & 51.60 & $-1.10$ \\
    All Delta & 45.0 & 57.1 & 51.05 & $-1.65$ \\
    \bottomrule
    \end{tabularx}
    \end{minipage}\hfill
    \begin{minipage}[t]{0.48\textwidth}
    \caption{Ablation on audio compression strategy. We compare our audio compression method against mainstream compression alternatives. Qwen2.5-Omni-3B with ($\rho_v{=}0.6,\;\rho_a{=}0.25$).}
    \label{tab:ablation-audio}
    \centering
    \scriptsize
    \renewcommand{\arraystretch}{1.18}
    \begin{tabularx}{\linewidth}{Xcccc}
    \toprule
    Variant & WS & DO & Avg. & $\Delta$ \\
    \midrule
    \textbf{Ours (full)} & \textbf{46.1} & \textbf{59.3} & \textbf{52.70} & -- \\
    \midrule
    Uniform sampling & 45.7 & 59.2 & 52.45 & $-0.25$ \\
    Average pooling & 45.4 & 58.2 & 51.80 & $-0.90$ \\
    Random drop & 45.3 & 58.2 & 51.75 & $-0.95$ \\
    Energy-based & 44.1 & 55.8 & 49.95 & $-2.75$ \\
    \bottomrule
    \end{tabularx}
    \end{minipage}
    \end{table*}
    \vspace{-1.2em}
\paragraph{Ablation Study of Visual Pruning Strategy.}
As shown in Table~\ref{tab:ablation-vision}, both the \textit{All Anchor} and \textit{All Delta} variants lead to noticeable performance degradation, demonstrating that simultaneously preserving temporal dynamics and spatial structural information is crucial for video understanding. Under this alternating framework, applying dynamic compression to Delta frames consistently outperforms the fixed compression strategy. This indicates the necessity of an optimal trade-off between incremental information density and global semantic information across varying compression rates. 
Furthermore, we conduct a sensitivity analysis on the switching threshold $\tau_r$ evaluated on the DailyOmni dataset. As illustrated in Appendix~E, the performance curve exhibits a clear inverted U-shaped trend, reaching its optimal peak in the middle region. This demonstrates that our method possesses excellent robustness regarding threshold selection.


\paragraph{Ablation Study of Audio Compression Strategy.}
As shown in Table~\ref{tab:ablation-audio}, we compare our method against four mainstream alternatives. Methods that directly discard tokens (random drop, energy-based filtering) suffer the largest degradation, as they break temporal continuity and irreversibly lose information; energy-based filtering performs worst because token energy in the audio embedding space reflects acoustic loudness rather than semantic importance. Uniform sampling and average pooling preserve temporal structure but do not fully exploit the information carried by the compressed tokens. Our method merges similar tokens rather than discarding them, maximally preserving information while maintaining global temporal ordering, and therefore achieves the best performance.

\section{Conclusion}
We present OmniScope, a training-free, modality-decoupled token compression framework for omnimodal large language models. OmniScope uses the query as a shared semantic anchor while allowing visual and audio tokens to independently assess their importance and allocate compression budgets, avoiding the fragility of unidirectional guidance under cross-modal salience mismatch. On top of this decoupled allocation, the visual side employs anchor-delta spatio-temporal compression and the audio side adopts per-second token merging, each minimizing information loss for its respective modality. On four audio-video understanding benchmarks at two model scales, OmniScope achieves the best average performance at all compression ratios, while delivering substantial prefilling speedup and memory savings.




\clearpage
\bibliographystyle{plainnat}
\bibliography{samples/ref}

\begin{thebibliography}{54}
\providecommand{\natexlab}[1]{#1}
\providecommand{\url}[1]{\texttt{#1}}
\expandafter\ifx\csname urlstyle\endcsname\relax
  \providecommand{\doi}[1]{doi: #1}\else
  \providecommand{\doi}{doi: \begingroup \urlstyle{rm}\Url}\fi

\bibitem[Bolya et~al.(2022)Bolya, Fu, Dai, Zhang, Feichtenhofer, and
  Hoffman]{bolya2022token}
Daniel Bolya, Cheng-Yang Fu, Xiaoliang Dai, Peizhao Zhang, Christoph
  Feichtenhofer, and Judy Hoffman.
\newblock Token merging: Your vit but faster.
\newblock \emph{arXiv preprint arXiv:2210.09461}, 2022.

\bibitem[Chen et~al.(2024{\natexlab{a}})Chen, Zhao, Liu, Bai, Lin, Zhou, and
  Chang]{chen2024image}
Liang Chen, Haozhe Zhao, Tianyu Liu, Shuai Bai, Junyang Lin, Chang Zhou, and
  Baobao Chang.
\newblock An image is worth 1/2 tokens after layer 2: Plug-and-play inference
  acceleration for large vision-language models.
\newblock In \emph{European Conference on Computer Vision}, pages 19--35.
  Springer, 2024{\natexlab{a}}.

\bibitem[Chen et~al.(2024{\natexlab{b}})Chen, Wang, Tian, Ye, Gao, Cui, Tong,
  Hu, Luo, Ma, et~al.]{chen2024far}
Zhe Chen, Weiyun Wang, Hao Tian, Shenglong Ye, Zhangwei Gao, Erfei Cui, Wenwen
  Tong, Kongzhi Hu, Jiapeng Luo, Zheng Ma, et~al.
\newblock How far are we to gpt-4v? closing the gap to commercial multimodal
  models with open-source suites.
\newblock \emph{Science China Information Sciences}, 67\penalty0 (12):\penalty0
  220101, 2024{\natexlab{b}}.

\bibitem[Dai et~al.(2023)Dai, Li, Li, Tiong, Zhao, Wang, Li, Fung, and
  Hoi]{dai2023instructblip}
Wenliang Dai, Junnan Li, Dongxu Li, Anthony Tiong, Junqi Zhao, Weisheng Wang,
  Boyang Li, Pascale~N Fung, and Steven Hoi.
\newblock Instructblip: Towards general-purpose vision-language models with
  instruction tuning.
\newblock \emph{Advances in neural information processing systems},
  36:\penalty0 49250--49267, 2023.

\bibitem[Dao(2023)]{dao2023flashattention}
Tri Dao.
\newblock Flashattention-2: Faster attention with better parallelism and work
  partitioning.
\newblock \emph{arXiv preprint arXiv:2307.08691}, 2023.

\bibitem[Ding et~al.(2026)Ding, Ji, Li, Liu, Chen, Wu, Li, Zeng, Shi, Guan,
  et~al.]{ding2026omnisift}
Yue Ding, Yiyan Ji, Jungang Li, Xuyang Liu, Xinlong Chen, Junfei Wu, Bozhou Li,
  Bohan Zeng, Yang Shi, Yushuo Guan, et~al.
\newblock Omnisift: Modality-asymmetric token compression for efficient
  omni-modal large language models.
\newblock \emph{arXiv preprint arXiv:2602.04804}, 2026.

\bibitem[Du et~al.(2016)Du, Ding, and Jia]{du2016study}
Mingjing Du, Shifei Ding, and Hongjie Jia.
\newblock Study on density peaks clustering based on k-nearest neighbors and
  principal component analysis.
\newblock \emph{Knowledge-Based Systems}, 99:\penalty0 135--145, 2016.

\bibitem[Frantar and Alistarh(2023)]{frantar2023sparsegpt}
Elias Frantar and Dan Alistarh.
\newblock Sparsegpt: Massive language models can be accurately pruned in
  one-shot.
\newblock In \emph{International conference on machine learning}, pages
  10323--10337. PMLR, 2023.

\bibitem[Fu et~al.(2025{\natexlab{a}})Fu, Dai, Luo, Li, Ren, Zhang, Wang, Zhou,
  Shen, Zhang, et~al.]{fu2025video}
Chaoyou Fu, Yuhan Dai, Yongdong Luo, Lei Li, Shuhuai Ren, Renrui Zhang, Zihan
  Wang, Chenyu Zhou, Yunhang Shen, Mengdan Zhang, et~al.
\newblock Video-mme: The first-ever comprehensive evaluation benchmark of
  multi-modal llms in video analysis.
\newblock In \emph{Proceedings of the IEEE/CVF conference on computer vision
  and pattern recognition}, pages 24108--24118, 2025{\natexlab{a}}.

\bibitem[Fu et~al.(2025{\natexlab{b}})Fu, Lin, Wang, Zhang, Shen, Liu, Cao,
  Long, Gao, Li, et~al.]{fu2025vita}
Chaoyou Fu, Haojia Lin, Xiong Wang, Yi-Fan Zhang, Yunhang Shen, Xiaoyu Liu,
  Haoyu Cao, Zuwei Long, Heting Gao, Ke~Li, et~al.
\newblock Vita-1.5: Towards gpt-4o level real-time vision and speech
  interaction.
\newblock \emph{arXiv preprint arXiv:2501.01957}, 2025{\natexlab{b}}.

\bibitem[Hong et~al.(2025)Hong, Yan, Cai, Jiang, Hu, and
  Xie]{hong2025worldsense}
Jack Hong, Shilin Yan, Jiayin Cai, Xiaolong Jiang, Yao Hu, and Weidi Xie.
\newblock Worldsense: Evaluating real-world omnimodal understanding for
  multimodal llms.
\newblock \emph{arXiv preprint arXiv:2502.04326}, 2025.

\bibitem[Huang et~al.(2025)Huang, Zhou, and Han]{huang2025prunevid}
Xiaohu Huang, Hao Zhou, and Kai Han.
\newblock Prunevid: Visual token pruning for efficient video large language
  models.
\newblock In \emph{Findings of the Association for Computational Linguistics:
  ACL 2025}, pages 19959--19973, 2025.

\bibitem[Hurst et~al.(2024)Hurst, Lerer, Goucher, Perelman, Ramesh, Clark,
  Ostrow, Welihinda, Hayes, Radford, et~al.]{hurst2024gpt}
Aaron Hurst, Adam Lerer, Adam~P Goucher, Adam Perelman, Aditya Ramesh, Aidan
  Clark, AJ~Ostrow, Akila Welihinda, Alan Hayes, Alec Radford, et~al.
\newblock Gpt-4o system card.
\newblock \emph{arXiv preprint arXiv:2410.21276}, 2024.

\bibitem[Jin et~al.(2025)Jin, Li, Gu, Liu, Zhao, Lai, Gan, Wang, Wang, Tan,
  et~al.]{jin2025efficient}
Yizhang Jin, Jian Li, Tianjun Gu, Yexin Liu, Bo~Zhao, Jinxiang Lai, Zhenye Gan,
  Yabiao Wang, Chengjie Wang, Xin Tan, et~al.
\newblock Efficient multimodal large language models: A survey.
\newblock \emph{Visual Intelligence}, 3\penalty0 (1):\penalty0 27, 2025.

\bibitem[Lee and Lee(2025)]{lee2025token}
Taehan Lee and Hyukjun Lee.
\newblock Token pruning in audio transformers: Optimizing performance and
  decoding patch importance.
\newblock \emph{arXiv preprint arXiv:2504.01690}, 2025.

\bibitem[Li et~al.(2024)Li, Zhang, Guo, Zhang, Li, Zhang, Zhang, Zhang, Li,
  Liu, et~al.]{li2024llava}
Bo~Li, Yuanhan Zhang, Dong Guo, Renrui Zhang, Feng Li, Hao Zhang, Kaichen
  Zhang, Peiyuan Zhang, Yanwei Li, Ziwei Liu, et~al.
\newblock Llava-onevision: Easy visual task transfer.
\newblock \emph{arXiv preprint arXiv:2408.03326}, 2024.

\bibitem[Li et~al.(2025)Li, Chen, Ji, Xu, Cui, Li, Zhang, Wang, Song, Zhang,
  et~al.]{li2025omnivideobench}
Caorui Li, Yu~Chen, Yiyan Ji, Jin Xu, Zhenyu Cui, Shihao Li, Yuanxing Zhang,
  Wentao Wang, Zhenghao Song, Dingling Zhang, et~al.
\newblock Omnivideobench: Towards audio-visual understanding evaluation for
  omni mllms.
\newblock \emph{arXiv preprint arXiv:2510.10689}, 2025.

\bibitem[Li et~al.(2023)Li, Wu, Li, and Liu]{li2023accelerating}
Yuang Li, Yu~Wu, Jinyu Li, and Shujie Liu.
\newblock Accelerating transducers through adjacent token merging.
\newblock \emph{arXiv preprint arXiv:2306.16009}, 2023.

\bibitem[Lin et~al.(2024)Lin, Ye, Zhu, Cui, Ning, Jin, and Yuan]{lin2024video}
Bin Lin, Yang Ye, Bin Zhu, Jiaxi Cui, Munan Ning, Peng Jin, and Li~Yuan.
\newblock Video-llava: Learning united visual representation by alignment
  before projection.
\newblock In \emph{Proceedings of the 2024 conference on empirical methods in
  natural language processing}, pages 5971--5984, 2024.

\bibitem[Lin et~al.(2025)Lin, Fu, Zhang, Liu, Zhang, Sun, Li, and
  Chen]{lin2025speechprune}
Yueqian Lin, Yuzhe Fu, Jingyang Zhang, Yudong Liu, Jianyi Zhang, Jingwei Sun,
  Hai~Helen Li, and Yiran Chen.
\newblock Speechprune: Context-aware token pruning for speech information
  retrieval.
\newblock In \emph{2025 IEEE International Conference on Multimedia and Expo
  (ICME)}, pages 1--6. IEEE, 2025.

\bibitem[Liu et~al.(2023)Liu, Li, Wu, and Lee]{liu2023visual}
Haotian Liu, Chunyuan Li, Qingyang Wu, and Yong~Jae Lee.
\newblock Visual instruction tuning.
\newblock \emph{Advances in neural information processing systems},
  36:\penalty0 34892--34916, 2023.

\bibitem[Liu et~al.(2025)Liu, Li, Sun, Wu, Gao, Zhang, Zhang, Jin, Yu, Zhan,
  et~al.]{liu2025javisgpt}
Kai Liu, Jungang Li, Yuchong Sun, Shengqiong Wu, Jianzhang Gao, Daoan Zhang,
  Wei Zhang, Sheng Jin, Sicheng Yu, Geng Zhan, et~al.
\newblock Javisgpt: A unified multi-modal llm for sounding-video comprehension
  and generation.
\newblock \emph{arXiv preprint arXiv:2512.22905}, 2025.

\bibitem[Luo et~al.(2026)Luo, Chen, Huang, Yin, Lin, Huang, Fu, Ji, Zheng, and
  Luo]{luo2026quota}
Yongdong Luo, Wang Chen, Weizhong Huang, Shukang Yin, Haojia Lin, Jinfa Huang,
  Chaoyou Fu, Jiayi Ji, Xiawu Zheng, and Jiebo Luo.
\newblock Quota: Query-oriented token assignment via cot query decouple for
  long video comprehension.
\newblock In \emph{Proceedings of the AAAI Conference on Artificial
  Intelligence}, volume~40, pages 24160--24168, 2026.

\bibitem[Ma et~al.(2023{\natexlab{a}})Ma, Fang, and Wang]{ma2023llm}
Xinyin Ma, Gongfan Fang, and Xinchao Wang.
\newblock Llm-pruner: On the structural pruning of large language models.
\newblock \emph{Advances in neural information processing systems},
  36:\penalty0 21702--21720, 2023{\natexlab{a}}.

\bibitem[Ma et~al.(2023{\natexlab{b}})Ma, Jin, Zheng, Wang, Li, Wu, Jiang,
  Zhang, and Ji]{ma2023ompq}
Yuexiao Ma, Taisong Jin, Xiawu Zheng, Yan Wang, Huixia Li, Yongjian Wu, Guannan
  Jiang, Wei Zhang, and Rongrong Ji.
\newblock Ompq: Orthogonal mixed precision quantization.
\newblock In \emph{Proceedings of the AAAI conference on artificial
  intelligence}, volume~37, pages 9029--9037, 2023{\natexlab{b}}.

\bibitem[Ma et~al.(2024)Ma, Li, Zheng, Ling, Xiao, Wang, Wen, Chao, and
  Ji]{ma2024affinequant}
Yuexiao Ma, Huixia Li, Xiawu Zheng, Feng Ling, Xuefeng Xiao, Rui Wang, Shilei
  Wen, Fei Chao, and Rongrong Ji.
\newblock Affinequant: Affine transformation quantization for large language
  models.
\newblock In \emph{International Conference on Learning Representations},
  volume 2024, pages 50932--50951, 2024.

\bibitem[Oyama et~al.(2023)Oyama, Yokoi, and Shimodaira]{oyama2023norm}
Momose Oyama, Sho Yokoi, and Hidetoshi Shimodaira.
\newblock Norm of word embedding encodes information gain.
\newblock In \emph{Proceedings of the 2023 Conference on Empirical Methods in
  Natural Language Processing}, pages 2108--2130, 2023.

\bibitem[Rabiner and Juang(1993)]{rabiner1993fundamentals}
Lawrence Rabiner and Biing-Hwang Juang.
\newblock \emph{Fundamentals of speech recognition}.
\newblock Prentice-Hall, Inc., 1993.

\bibitem[Radford et~al.(2021)Radford, Kim, Hallacy, Ramesh, Goh, Agarwal,
  Sastry, Askell, Mishkin, Clark, et~al.]{radford2021learning}
Alec Radford, Jong~Wook Kim, Chris Hallacy, Aditya Ramesh, Gabriel Goh,
  Sandhini Agarwal, Girish Sastry, Amanda Askell, Pamela Mishkin, Jack Clark,
  et~al.
\newblock Learning transferable visual models from natural language
  supervision.
\newblock In \emph{International conference on machine learning}, pages
  8748--8763. PmLR, 2021.

\bibitem[Shang et~al.(2025)Shang, Cai, Xu, Lee, and Yan]{shang2025llava}
Yuzhang Shang, Mu~Cai, Bingxin Xu, Yong~Jae Lee, and Yan Yan.
\newblock Llava-prumerge: Adaptive token reduction for efficient large
  multimodal models.
\newblock In \emph{Proceedings of the IEEE/CVF International Conference on
  Computer Vision}, pages 22857--22867, 2025.

\bibitem[Shao et~al.(2025{\natexlab{a}})Shao, Tao, Qin, You, Sui, and
  Wang]{shao2025holitom}
Kele Shao, Keda Tao, Can Qin, Haoxuan You, Yang Sui, and Huan Wang.
\newblock Holitom: Holistic token merging for fast video large language models.
\newblock \emph{arXiv preprint arXiv:2505.21334}, 2025{\natexlab{a}}.

\bibitem[Shao et~al.(2025{\natexlab{b}})Shao, Tao, Zhang, Feng, Cai, Shang,
  You, Qin, Sui, and Wang]{shao2025tokens}
Kele Shao, Keda Tao, Kejia Zhang, Sicheng Feng, Mu~Cai, Yuzhang Shang, Haoxuan
  You, Can Qin, Yang Sui, and Huan Wang.
\newblock When tokens talk too much: A survey of multimodal long-context token
  compression across images, videos, and audios.
\newblock \emph{arXiv preprint arXiv:2507.20198}, 2025{\natexlab{b}}.

\bibitem[Song et~al.(2024)Song, Wang, Chen, Wang, Guan, and
  Wang]{song2024moresimpleeffectivetoken}
Dingjie Song, Wenjun Wang, Shunian Chen, Xidong Wang, Michael Guan, and Benyou
  Wang.
\newblock Less is more: A simple yet effective token reduction method for
  efficient multi-modal llms, 2024.

\bibitem[Sun et~al.(2024)Sun, Yu, Tang, Chen, Tan, Li, Lu, Ma, Wang, and
  Zhang]{sun2024video}
Guangzhi Sun, Wenyi Yu, Changli Tang, Xianzhao Chen, Tian Tan, Wei Li, Lu~Lu,
  Zejun Ma, Yuxuan Wang, and Chao Zhang.
\newblock video-salmonn: Speech-enhanced audio-visual large language models.
\newblock \emph{arXiv preprint arXiv:2406.15704}, 2024.

\bibitem[Sun et~al.(2025)Sun, Xin, Li, Sun, Lin, and
  Batista-Navarro]{sun2025lvpruning}
Yizheng Sun, Yanze Xin, Hao Li, Jingyuan Sun, Chenghua Lin, and Riza~Theresa
  Batista-Navarro.
\newblock Lvpruning: An effective yet simple language-guided vision token
  pruning approach for multi-modal large language models.
\newblock In \emph{Findings of the Association for Computational Linguistics:
  NAACL 2025}, pages 4299--4308, 2025.

\bibitem[Tang et~al.(2025)Tang, Li, Yang, Zhuang, Sun, Li, Ma, and
  Zhang]{tang2025video}
Changli Tang, Yixuan Li, Yudong Yang, Jimin Zhuang, Guangzhi Sun, Wei Li, Zejun
  Ma, and Chao Zhang.
\newblock video-salmonn 2: Caption-enhanced audio-visual large language models.
\newblock \emph{arXiv preprint arXiv:2506.15220}, 2025.

\bibitem[Tao et~al.(2025{\natexlab{a}})Tao, Qin, You, Sui, and
  Wang]{tao2025dycoke}
Keda Tao, Can Qin, Haoxuan You, Yang Sui, and Huan Wang.
\newblock Dycoke: Dynamic compression of tokens for fast video large language
  models.
\newblock In \emph{Proceedings of the Computer Vision and Pattern Recognition
  Conference}, pages 18992--19001, 2025{\natexlab{a}}.

\bibitem[Tao et~al.(2025{\natexlab{b}})Tao, Shao, Yu, Wang, Wang,
  et~al.]{tao2025omnizip}
Keda Tao, Kele Shao, Bohan Yu, Weiqiang Wang, Huan Wang, et~al.
\newblock Omnizip: Audio-guided dynamic token compression for fast omnimodal
  large language models.
\newblock \emph{arXiv preprint arXiv:2511.14582}, 2025{\natexlab{b}}.

\bibitem[Team et~al.(2023)Team, Anil, Borgeaud, Alayrac, Yu, Soricut,
  Schalkwyk, Dai, Hauth, Millican, et~al.]{team2023gemini}
Gemini Team, Rohan Anil, Sebastian Borgeaud, Jean-Baptiste Alayrac, Jiahui Yu,
  Radu Soricut, Johan Schalkwyk, Andrew~M Dai, Anja Hauth, Katie Millican,
  et~al.
\newblock Gemini: a family of highly capable multimodal models.
\newblock \emph{arXiv preprint arXiv:2312.11805}, 2023.

\bibitem[Wang et~al.(2024)Wang, Bai, Tan, Wang, Fan, Bai, Chen, Liu, Wang, Ge,
  et~al.]{wang2024qwen2}
Peng Wang, Shuai Bai, Sinan Tan, Shijie Wang, Zhihao Fan, Jinze Bai, Keqin
  Chen, Xuejing Liu, Jialin Wang, Wenbin Ge, et~al.
\newblock Qwen2-vl: Enhancing vision-language model's perception of the world
  at any resolution.
\newblock \emph{arXiv preprint arXiv:2409.12191}, 2024.

\bibitem[Weng et~al.(2024)Weng, Han, He, Chang, and Zhuang]{weng2024longvlm}
Yuetian Weng, Mingfei Han, Haoyu He, Xiaojun Chang, and Bohan Zhuang.
\newblock Longvlm: Efficient long video understanding via large language
  models.
\newblock In \emph{European Conference on Computer Vision}, pages 453--470.
  Springer, 2024.

\bibitem[Wu et~al.(2023)Wu, Chen, Zhang, Hui, Berg-Kirkpatrick, and
  Dubnov]{wu2023large}
Yusong Wu, Ke~Chen, Tianyu Zhang, Yuchen Hui, Taylor Berg-Kirkpatrick, and
  Shlomo Dubnov.
\newblock Large-scale contrastive language-audio pretraining with feature
  fusion and keyword-to-caption augmentation.
\newblock In \emph{ICASSP 2023-2023 IEEE International Conference on Acoustics,
  Speech and Signal Processing (ICASSP)}, pages 1--5. IEEE, 2023.

\bibitem[Xiao et~al.(2023)Xiao, Lin, Seznec, Wu, Demouth, and
  Han]{xiao2023smoothquant}
Guangxuan Xiao, Ji~Lin, Mickael Seznec, Hao Wu, Julien Demouth, and Song Han.
\newblock Smoothquant: Accurate and efficient post-training quantization for
  large language models.
\newblock In \emph{International conference on machine learning}, pages
  38087--38099. PMLR, 2023.

\bibitem[Xie and Wu(2024)]{xie2024mini}
Zhifei Xie and Changqiao Wu.
\newblock Mini-omni2: Towards open-source gpt-4o with vision, speech and duplex
  capabilities.
\newblock \emph{arXiv preprint arXiv:2410.11190}, 2024.

\bibitem[Xing et~al.(2024)Xing, Huang, Dong, Lu, Zhang, Zang, Cao, He, Wang,
  Wu, et~al.]{xing2024pyramiddrop}
Long Xing, Qidong Huang, Xiaoyi Dong, Jiajie Lu, Pan Zhang, Yuhang Zang, Yuhang
  Cao, Conghui He, Jiaqi Wang, Feng Wu, et~al.
\newblock Pyramiddrop: Accelerating your large vision-language models via
  pyramid visual redundancy reduction.
\newblock \emph{arXiv preprint arXiv:2410.17247}, 2024.

\bibitem[Xu et~al.(2025{\natexlab{a}})Xu, Guo, He, Hu, He, Bai, Chen, Wang,
  Fan, Dang, et~al.]{xu2025qwen2}
Jin Xu, Zhifang Guo, Jinzheng He, Hangrui Hu, Ting He, Shuai Bai, Keqin Chen,
  Jialin Wang, Yang Fan, Kai Dang, et~al.
\newblock Qwen2.5-omni technical report.
\newblock \emph{arXiv preprint arXiv:2503.20215}, 2025{\natexlab{a}}.

\bibitem[Xu et~al.(2025{\natexlab{b}})Xu, Guo, Hu, Chu, Wang, He, Wang, Shi,
  He, Zhu, et~al.]{xu2025qwen3}
Jin Xu, Zhifang Guo, Hangrui Hu, Yunfei Chu, Xiong Wang, Jinzheng He, Yuxuan
  Wang, Xian Shi, Ting He, Xinfa Zhu, et~al.
\newblock Qwen3-omni technical report.
\newblock \emph{arXiv preprint arXiv:2509.17765}, 2025{\natexlab{b}}.

\bibitem[Yang et~al.(2025{\natexlab{a}})Yang, Yao, Chen, Fu, Bai, Zhao, Sun,
  Yin, Wei, and Zhou]{yang2025humanomniv2}
Qize Yang, Shimin Yao, Weixuan Chen, Shenghao Fu, Detao Bai, Jiaxing Zhao,
  Boyuan Sun, Bowen Yin, Xihan Wei, and Jingren Zhou.
\newblock Humanomniv2: From understanding to omni-modal reasoning with context.
\newblock \emph{arXiv preprint arXiv:2506.21277}, 2025{\natexlab{a}}.

\bibitem[Yang et~al.(2025{\natexlab{b}})Yang, Chen, Tian, Wang, Li, Yu, and
  Jia]{yang2025visionzip}
Senqiao Yang, Yukang Chen, Zhuotao Tian, Chengyao Wang, Jingyao Li, Bei Yu, and
  Jiaya Jia.
\newblock Visionzip: Longer is better but not necessary in vision language
  models.
\newblock In \emph{Proceedings of the IEEE/CVF Conference on Computer Vision
  and Pattern Recognition}, pages 19792--19802, 2025{\natexlab{b}}.

\bibitem[Ye et~al.(2025)Ye, Yang, Goel, Huang, Zhu, Su, Lin, Cheng, Wan, Tian,
  et~al.]{ye2025omnivinci}
Hanrong Ye, Chao-Han~Huck Yang, Arushi Goel, Wei Huang, Ligeng Zhu, Yuanhang
  Su, Sean Lin, An-Chieh Cheng, Zhen Wan, Jinchuan Tian, et~al.
\newblock Omnivinci: Enhancing architecture and data for omni-modal
  understanding llm.
\newblock \emph{arXiv preprint arXiv:2510.15870}, 2025.

\bibitem[Zhang et~al.(2024)Zhang, Fan, Ma, Zheng, Huang, Cheng, Gudovskiy,
  Okuno, Nakata, Keutzer, et~al.]{zhang2024sparsevlm}
Yuan Zhang, Chun-Kai Fan, Junpeng Ma, Wenzhao Zheng, Tao Huang, Kuan Cheng,
  Denis Gudovskiy, Tomoyuki Okuno, Yohei Nakata, Kurt Keutzer, et~al.
\newblock Sparsevlm: Visual token sparsification for efficient vision-language
  model inference.
\newblock \emph{arXiv preprint arXiv:2410.04417}, 2024.

\bibitem[Zhao et~al.(2025)Zhao, Wei, and Bo]{zhao2025r1}
Jiaxing Zhao, Xihan Wei, and Liefeng Bo.
\newblock R1-omni: Explainable omni-multimodal emotion recognition with
  reinforcement learning.
\newblock \emph{arXiv preprint arXiv:2503.05379}, 2025.

\bibitem[Zheng et~al.(2021)Zheng, Ma, Xi, Zhang, Ding, Li, Chen, Tian, and
  Ji]{zheng2021information}
Xiawu Zheng, Yuexiao Ma, Teng Xi, Gang Zhang, Errui Ding, Yuchao Li, Jie Chen,
  Yonghong Tian, and Rongrong Ji.
\newblock An information theory-inspired strategy for automatic network
  pruning.
\newblock \emph{arXiv preprint arXiv:2108.08532}, 2021.

\bibitem[Zhou et~al.(2025)Zhou, Wang, and Wu]{zhou2025daily}
Ziwei Zhou, Rui Wang, and Zuxuan Wu.
\newblock Daily-omni: Towards audio-visual reasoning with temporal alignment
  across modalities.
\newblock \emph{arXiv preprint arXiv:2505.17862}, 2025.

\end{thebibliography}

\clearpage
\beginappendix
\appendix
\setcounter{table}{5}   
\setcounter{figure}{4}  
\section{Per-Domain Results on WorldSense}
\label{app:worldsense}
WorldSense spans eight diverse domains. As shown in Table~\ref{tab:worldsense}, OmniScope achieves the highest average accuracy at both 45\% and 25\% retention across both model scales, and remains competitive across individual domains. In contrast, competing methods exhibit uneven performance across domains---for instance, on the 7B model, OmniZip performs well on \emph{Culture \& Politics} but drops sharply on \emph{Performance} and \emph{Games} at 25\% retention. This further validates that independently compressing each modality based on query relevance provides consistent robustness across diverse audio-visual scenarios.

\begin{table}[H]
\caption{Per-domain results on WorldSense. The \textbf{best} result among compression methods for each metric is bolded.}
\label{tab:worldsense}
\centering
\resizebox{\textwidth}{!}{%
\begin{tabular}{llccccccccc}
\toprule
Method & Ratio & Tech \& Sci. & Cul. \& Pol. & Daily Life & Film \& TV & Perform. & Games & Sports & Music & Avg \\
\midrule
\multicolumn{11}{c}{\emph{Qwen2.5-Omni-7B}} \\
\midrule
Full Tokens & 100\% & 48.0 & 48.9 & 45.0 & 47.0 & 46.1 & 40.3 & 44.0 & 47.3 & 46.0 \\
\midrule
Random             & 45\% & 48.8 & 49.2 & 43.9 & 42.5 & 43.1 & 39.9 & 42.6 & 45.8 & 44.7 \\
FastV (A\&V)       & 45\% & 47.6 & 49.5 & 46.4 & 43.5 & 46.1 & \textbf{41.2} & \textbf{44.2} & 46.6 & 45.8 \\
OmniZip            & 45\% & \textbf{49.4} & \textbf{50.2} & 45.9 & 44.1 & 43.1 & 39.9 & 43.3 & \textbf{46.8} & 45.7 \\
\textbf{OmniScope (Ours)} & 45\% & 48.6 & \textbf{50.2} & \textbf{46.8} & \textbf{45.1} & \textbf{47.6} & 40.3 & 43.7 & 46.1 & \textbf{46.3} \\
\midrule
Random             & 25\% & \textbf{48.4} & 47.2 & 44.2 & \textbf{44.3} & 42.7 & 41.2 & 42.1 & \textbf{47.0} & 44.9 \\
FastV (A\&V)       & 25\% & 46.9 & 48.5 & 45.9 & 42.7 & \textbf{44.9} & \textbf{41.6} & 40.7 & 44.8 & 44.7 \\
OmniZip            & 25\% & 46.9 & \textbf{50.5} & 45.4 & 42.7 & 41.9 & 38.2 & 42.3 & 46.8 & 44.8 \\
\textbf{OmniScope (Ours)} & 25\% & \textbf{48.4} & 49.5 & \textbf{46.2} & \textbf{44.3} & \textbf{44.9} & 38.6 & \textbf{43.5} & \textbf{47.0} & \textbf{45.7} \\
\midrule
\multicolumn{11}{c}{\emph{Qwen2.5-Omni-3B}} \\
\midrule
Full Tokens & 100\% & 52.7 & 49.8 & 43.6 & 46.2 & 41.6 & 42.5 & 44.2 & 46.3 & 46.1 \\
\midrule
Random             & 45\% & 50.6 & 45.0 & 44.5 & 42.7 & 40.4 & 40.3 & 40.7 & 44.6 & 44.1 \\
FastV (A\&V)       & 45\% & 51.4 & 46.9 & 44.4 & 43.0 & 39.3 & 40.3 & 40.2 & 44.1 & 44.2 \\
OmniZip            & 45\% & 52.2 & 46.6 & \textbf{45.1} & 44.9 & 40.4 & \textbf{42.5} & 41.6 & 46.1 & 45.4 \\
\textbf{OmniScope (Ours)} & 45\% & \textbf{52.4} & \textbf{47.6} & 45.0 & \textbf{45.9} & \textbf{42.7} & 41.2 & \textbf{43.3} & \textbf{47.0} & \textbf{46.1} \\
\midrule
Random             & 25\% & 48.8 & 44.0 & 40.6 & \textbf{43.5} & 40.8 & 36.1 & 40.0 & 43.6 & 42.5 \\
FastV (A\&V)       & 25\% & 49.6 & 43.0 & 42.9 & 43.3 & \textbf{41.6} & 35.6 & 39.3 & 43.6 & 42.9 \\
OmniZip            & 25\% & 49.8 & \textbf{47.6} & \textbf{43.8} & 42.5 & 40.8 & 39.9 & \textbf{40.2} & 44.6 & 44.0 \\
\textbf{OmniScope (Ours)} & 25\% & \textbf{52.9} & 46.9 & \textbf{43.8} & \textbf{43.5} & \textbf{41.6} & \textbf{40.3} & 38.8 & \textbf{45.1} & \textbf{44.5} \\
\bottomrule
\end{tabular}
}
\end{table}

\section{Hyperparameter Settings}
\label{app:hyperparams}
Table~\ref{tab:hyperparams} lists all hyperparameters used in OmniScope. The same configuration is used across all benchmarks and both model scales without per-dataset tuning. The audio scoring module includes the temperature coefficient, top-$k$ aggregation count, and neighbor boost parameters; the AD-STC module includes the switching threshold and the number of DPC-KNN neighbors.
\begin{table}[H]
\centering
\small
\begin{minipage}{0.56\textwidth}
\caption{Hyperparameter settings for OmniScope.}
\label{tab:hyperparams}
\begin{tabularx}{\linewidth}{lXl}
\toprule
Module & Parameter & Value \\
\midrule
Audio scoring & Temperature $\tau$ & 0.05 \\
 & Top-$k$ aggregation & 5 \\
 & Neighbor boost $\beta,\gamma,R,P$ & 1.5, 0.5, 3, 80 \\
AD-STC & Switching threshold $\tau_r$ & 0.4 \\
 & DPC-KNN neighbors $k$ & 5 \\
\bottomrule
\end{tabularx}
\end{minipage}
\end{table}

\section{CLIP Scoring Overhead Analysis}
\label{app:clip-overhead}

\begin{figure}[H]
  \centering
  \includegraphics[width=0.77\linewidth]{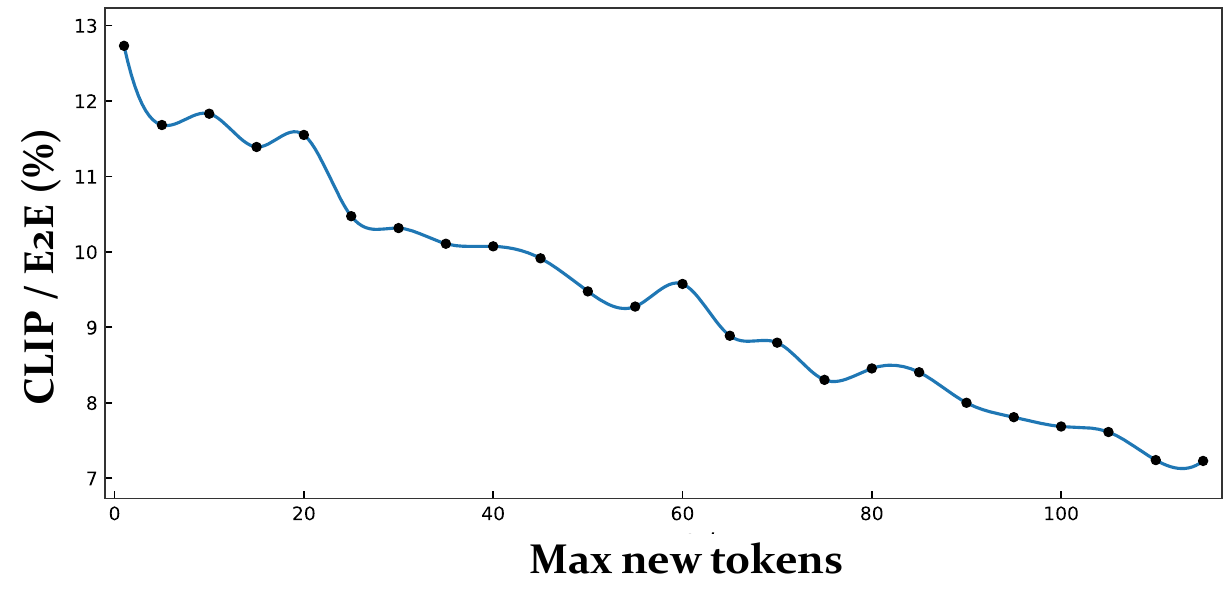}
  \caption{Ratio of CLIP visual scoring overhead to end-to-end latency as the number of generated tokens increases, measured on Qwen2.5-Omni-7B with 45\% retention.}
  \label{fig:clip-overhead}
  \Description{A line chart with the number of generated tokens on the horizontal axis and the ratio of CLIP visual scoring overhead to end-to-end latency on the vertical axis. The curve decreases steadily from approximately 13 percent when generating a single token to approximately 7 percent at 100 tokens, showing that the fixed CLIP scoring cost is progressively amortized as the generation length increases. Measured on Qwen2.5-Omni-7B with 45 percent token retention.}
\end{figure}

The CLIP visual scorer introduces a fixed one-time overhead. Fig.~\ref{fig:clip-overhead} shows that its share of end-to-end latency drops from approximately 13\% when generating a single token to around 7\% at 100 tokens, confirming that this cost is progressively amortized in longer generation scenarios.
\section{Visual Scoring Strategy Ablation}
\label{app:visual-scorer}

Table~\ref{tab:ablation-visual-scorer} compares the external CLIP scorer with the OmniLLM's own visual embeddings (Self-Embed) for frame-level importance estimation. CLIP scoring consistently achieves higher overall average accuracy across both model scales and compression ratios. Notably, the performance drop of Self-Embed on the 7B model is relatively small (approximately 0.5 points on average), whereas the drop on the 3B model is more pronounced (approximately 0.9 points on average), indicating that larger models already achieve substantially better vision--text alignment.

\begin{table}[H]
\caption{Ablation on visual scoring strategy. ``Self-Embed'' replaces the external CLIP scorer with the OmniLLM's own visual embeddings. $\Delta$ denotes the difference relative to the CLIP scoring variant. WS, DO, OVB, and VMME denote WorldSense, DailyOmni, OmniVideoBench, and Video-MME, respectively.}
\label{tab:ablation-visual-scorer}
\centering
\small
\begin{tabular}{llcccccr}
\toprule
Variant & Ratio & WS & DO & OVB & VMME & Avg. & $\Delta$ \\
\midrule
\multicolumn{8}{c}{\emph{Qwen2.5-Omni-7B}} \\
\midrule
\textbf{Ours (CLIP)} & 45\% & \textbf{46.3} & \textbf{60.5} & 35.5 & \textbf{64.3} & \textbf{51.65} & $-$ \\
Self-Embed            & 45\% & 45.8 & 59.1 & \textbf{35.8} & 64.0 & 51.18 & $-0.47$ \\
\textbf{Ours (CLIP)} & 25\% & \textbf{45.7} & \textbf{58.8} & \textbf{35.8} & \textbf{63.7} & \textbf{51.00} & $-$ \\
Self-Embed            & 25\% & 45.1 & 57.8 & \textbf{35.8} & 63.4 & 50.53 & $-0.47$ \\
\midrule
\multicolumn{8}{c}{\emph{Qwen2.5-Omni-3B}} \\
\midrule
\textbf{Ours (CLIP)} & 45\% & \textbf{46.1} & \textbf{59.3} & 33.2 & \textbf{62.1} & \textbf{50.18} & $-$ \\
Self-Embed            & 45\% & 45.1 & 57.2 & \textbf{33.4} & 61.3 & 49.25 & $-0.93$ \\
\textbf{Ours (CLIP)} & 25\% & \textbf{44.5} & \textbf{56.9} & \textbf{33.1} & \textbf{60.1} & \textbf{48.65} & $-$ \\
Self-Embed            & 25\% & 43.7 & 55.5 & 32.5 & 59.7 & 47.85 & $-0.80$ \\
\bottomrule
\end{tabular}
\end{table}

\section{Ratio Threshold Sensitivity Analysis}
\label{app:ratio-threshold}
\begin{figure}[H]
  \centering
  \includegraphics[width=0.78\linewidth]{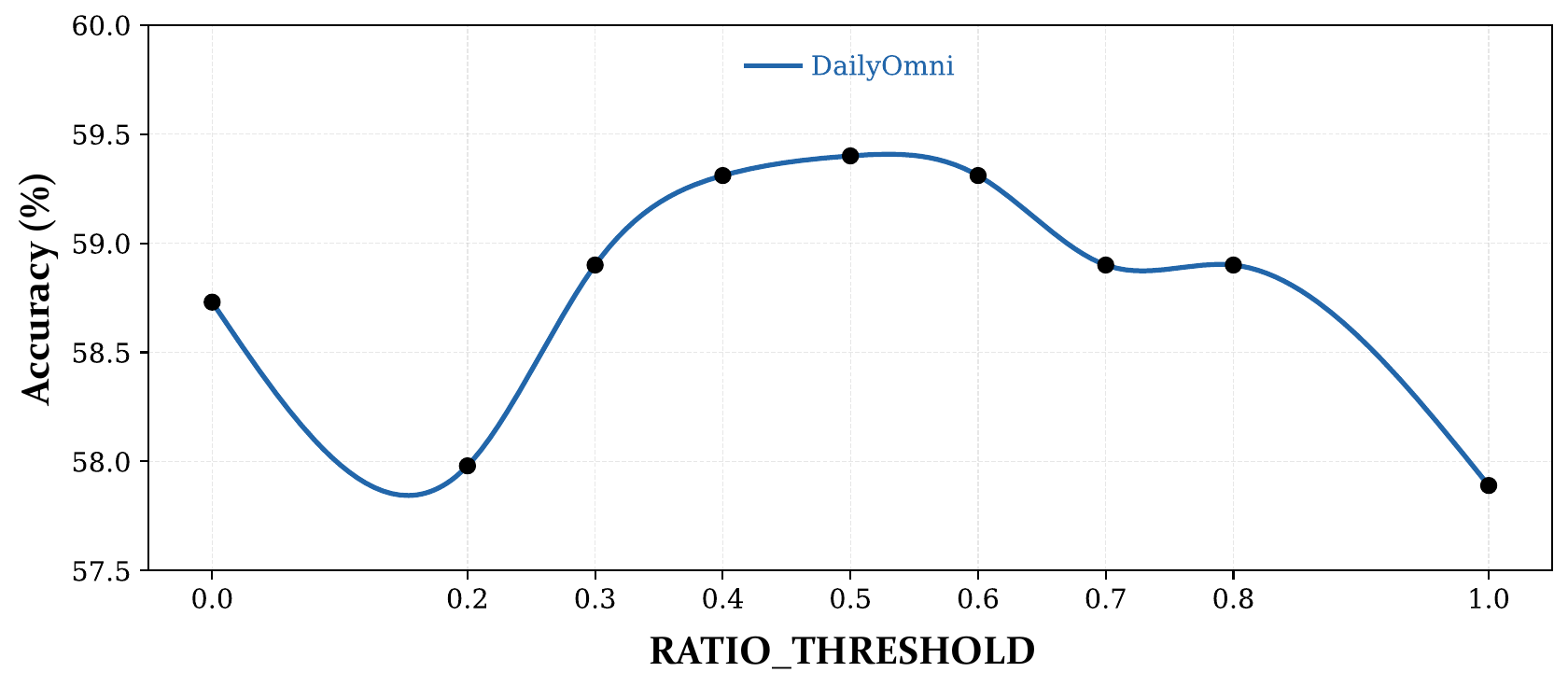}
  \caption{Effect of the ratio threshold $\tau_r$ on accuracy. $\tau_r$ controls the switching point between the two Delta frame scoring strategies in AD-STC. Evaluated on Qwen2.5-Omni-3B using the DailyOmni benchmark with ($\rho_v{=}0.6,\;\rho_a{=}0.25$).}
  \label{fig:ratio-threshold}
  \Description{A line chart with the switching threshold tau_r of the Delta frame scoring strategies in AD-STC on the horizontal axis and DailyOmni accuracy on the vertical axis. The curve exhibits a clear inverted U shape, reaching its optimal peak in the middle region and declining on both ends, indicating that the method is robust to threshold selection with a wide optimal range. Evaluated on Qwen2.5-Omni-3B with rho_v equals 0.6 and rho_a equals 0.25.}
\end{figure}

We conduct a sensitivity analysis on the switching threshold $\tau_r$ that controls the transition between the two Delta frame scoring strategies in AD-STC. As shown in Fig.~\ref{fig:ratio-threshold}, evaluated on the DailyOmni benchmark, the performance curve exhibits a clear inverted U-shaped trend, reaching its optimal peak in the middle region. This indicates that our method is robust to threshold selection, with a wide optimal range that does not require fine-grained tuning.

\section{Cross-Modal Salience Mismatch Analysis}
\label{app:salience-corr}
\begin{figure}[H]
  \centering
  \includegraphics[width=\linewidth]{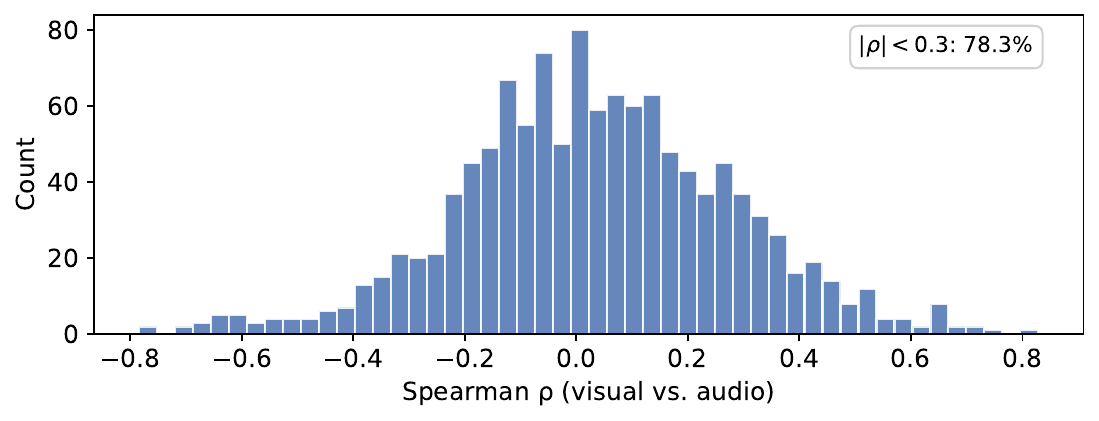}
  \caption{Distribution of Spearman correlation coefficients between query-conditioned visual (CLIP) and audio (CLAP) importance scores across 1,197 query-video pairs on DailyOmni. Approximately 78.3\% of samples exhibit only weak correlation ($|\rho| < 0.3$), confirming that cross-modal salience mismatch is pervasive rather than anecdotal.}
  \label{fig:salience-corr}
  \Description{A histogram with the Spearman correlation coefficient between query-conditioned visual (CLIP) and audio (CLAP) importance scores on the horizontal axis and the sample count on the vertical axis. The distribution is approximately symmetric and centered around zero; roughly 78.3 percent of samples have an absolute correlation coefficient below 0.3, falling in the weak-correlation range. This shows that the two modalities' relevance patterns for the same query are largely uncorrelated in the vast majority of cases, validating that cross-modal salience mismatch is pervasive. Data from 1,197 query-video pairs on the DailyOmni dataset.}
\end{figure}

To quantify the pervasiveness of cross-modal salience mismatch, we compute the Spearman correlation coefficient between query-conditioned visual and audio importance scores on the DailyOmni dataset. Visual scores are obtained via CLIP cosine similarity between each frame and the query, while audio scores are obtained via CLAP cosine similarity between each one-second segment and the query. Here, CLIP and CLAP serve as an external probe independent of the actual scoring pipeline in OmniScope (Sec.~3.3), so that the observation of this phenomenon does not depend on our method design. We compute the correlation pairwise across 1{,}197 query-video pairs, aligned at one-second granularity. As shown in Fig.~\ref{fig:salience-corr}, the distribution is centered around zero, with approximately 78.3\% of samples exhibiting only weak correlation ($|\rho| < 0.3$). This confirms that cross-modal salience mismatch is a pervasive phenomenon on this dataset.

\end{document}